\documentclass[preprint,12pt,authoryear]{elsarticle}

\makeatletter
\def\ps@pprintTitle{%
	\let\@oddhead\@empty
	\let\@evenhead\@empty
	\let\@oddfoot\@empty
	\let\@evenfoot\@oddfoot
}
\makeatother

\usepackage{amssymb}
\usepackage{amsmath}

\usepackage{siunitx}
\usepackage{graphicx}%
\usepackage{multirow}%
\usepackage{amsmath,amssymb,amsfonts}%
\usepackage{mathrsfs}%
\usepackage{xcolor}%
\usepackage{manyfoot}%
\usepackage{booktabs}%
\usepackage{listings}%
\usepackage{blkarray}
\usepackage{subcaption}
\usepackage{cleveref}

\usepackage{soul}
\usepackage{url}

\newcommand{\remove}[1]{}

\usepackage{tikz}
\usetikzlibrary{positioning}
\usetikzlibrary{arrows,plotmarks,decorations.markings,trees,shapes,pgfplots.groupplots,automata,circuits}
 \tikzset{dot/.style = {circle, fill, minimum size=#1,inner sep=0pt, outer sep=0pt, fill, circle},dot/.default = 6pt}
 \tikzset{dot2/.style = {circle, fill, color=black!40,minimum size=6pt,inner sep=0pt, outer sep=0pt, fill, circle}}
 \tikzset{dot3/.style = {circle, fill, color=blue!80,minimum size=6pt,inner sep=0pt, outer sep=0pt, fill, circle}}
 \tikzset{dot4/.style = {circle, fill, color=brown!80,minimum size=6pt,inner sep=0pt, outer sep=0pt, fill, circle}}
 \tikzstyle{a}=[->,>=stealth,dashed]
 \tikzstyle{a2}=[->,>=stealth]
 \tikzstyle{nodo}=[ellipse,draw=black!100,fill=black!0,line width=.7pt,minimum width=1.2cm,minimum height=0.8cm,text width=1.2cm,text centered]
 \tikzstyle{nodo2}=[ellipse,draw=black!100,fill=black!10,line width=.7pt,minimum width=1.2cm,minimum height=0.8cm,text width=1.2cm,text centered]
 \tikzstyle{nodo3}=[ellipse,draw=black!100,fill=black!30,line width=.7pt,minimum width=1.2cm,minimum height=0.8cm,text width=1.2cm,text centered]
 \tikzstyle{arco}=[draw=black!80,line width=.7pt, postaction={decorate}, decoration={markings,mark=at position 1.0 with {\arrow[ draw=black!80,line width=.7pt]{>}}}]

\usepackage{longtable}
\usepackage{adjustbox}
\usepackage{array}
\newcolumntype{N}{@{}m{0pt}@{}}
\newcolumntype{L}[1]{>{\raggedright\arraybackslash}p{#1}}
\newcolumntype{R}[1]{>{\raggedleft\arraybackslash}p{#1}}
\newcolumntype{C}[1]{>{\centering\arraybackslash}p{#1}}
\newcommand{\bmU}{\mathbf{U}}
\newcommand{\bmX}{\mathbf{X}}

\newcommand{\bmV}{\mathbf{V}}
\newcommand{\bmv}{\mathbf{v}}

\newcommand{\mcM}{\mathcal{M}}

\newcommand{\mcG}{\mathcal{G}}
\newcommand{\mcF}{\mathcal{F}}
\newcommand{\mcP}{\mathcal{P}}

\newtheorem{definition}{Definition}

\journal{}

\begin{document}

\begin{frontmatter}


\author[UALmat,UALcdtime]{Rafael Caba\~{n}as\corref{cor1}}\ead{rcabanas@ual.es}
\author[UALmat,UALcdtime]{Ana D. Maldonado}\ead{ana.d.maldonado@ual.es}
\author[UALmat,UALcdtime]{Mar\'{i}a Morales}\ead{maria.morales@ual.es}
\author[UALbio]{Pedro~A.~Aguilera}\ead{aguilera@ual.es}
\author[UALmat,UALcdtime]{Antonio Salmer\'{o}n}\ead{antonio.salmeron@ual.es}


\cortext[cor1]{Corresponding author}
\affiliation[UALmat]{organization={Department of Mathematics, University of Almería},
            addressline={Ctra. Sacramento s/n},
            city={La Cañada, Almería},
            postcode={04120},
            country={Spain}}
            
\affiliation[UALbio]{organization={Department of Biology and Geology, University of Almería},
            addressline={Ctra. Sacramento s/n},
            city={La Cañada, Almería},
            postcode={04120},
            country={Spain}}

\affiliation[UALcdtime]{organization={Center for the Development and Transfer of Matehmatical Research to Industry (CDTIME), University of Almería},
            addressline={Ctra. Sacramento s/n},
            city={La Cañada, Almería},
            postcode={04120},
            country={Spain}}
            

\title{Bayesian Networks for Causal Analysis in Socioecological Systems}

\begin{abstract}

Causal and counterfactual reasoning are emerging directions in data science that allow us to reason about hypothetical scenarios. This is particularly useful in fields like environmental and ecological sciences, where interventional data are usually not available.
Structural causal models are probabilistic models for causal analysis that simplify this kind of reasoning due to their graphical representation. They can be regarded as extensions of the so-called Bayesian networks, a well known modeling tool commonly used in environmental and ecological problems.
The main contribution of this paper is to analyze the relations of necessity and sufficiency between the variables of a socioecological system using counterfactual reasoning with Bayesian networks. In particular, we consider a case study involving socioeconomic factors and land-uses in southern Spain. 
In addition, this paper aims to be a coherent overview of the fundamental concepts for applying counterfactual reasoning, so that environmental researchers with a background in Bayesian networks can easily take advantage of the structural causal model formalism.

\end{abstract}




\end{frontmatter}



\section{Introduction}\label{sec:intro}
Causal (and counterfactual) reasoning \citep{pearl2009} allows to analyze cause-effect relationships, which is of fundamental importance for environmental and ecological practitioners and scientists. It can help in the development of effective strategies to mitigate or adapt to environmental problems, such as designing policies to reduce greenhouse gas emissions. 
This kind of reasoning can also help to evaluate the impact of different human activities on ecosystems.

Causal reasoning can typically be formally pursued through randomized experiments (a.k.a. randomized control trials), in which the variables of interest are intervened. In doing so, a study sample is randomly divided into one group that will receive the intervention with a given value and another that will be intervened with an alternative value. For example, in the problem of determining if a drug has a significant impact on the recovery from an illness, a group of patients will receive such drug whereas the other will receive a placebo. However, doing a randomized experiment in the field of environmental and ecological sciences might be expensive, unethical or directly impossible. For instance, if we aim to determine the influence of a population (from a specific species) on the structure and functioning of an ecosystem, it cannot be completely removed from it (or introduced in a new one where the population was not originally present). As a consequence, the environmental and ecological data available is usually observational, obtained from non-experimental studies. Using observational data with traditional statistical methods might lead to misleading conclusions when it comes to studying cause-effect relationships.

To illustrate the aforementioned problem, let us consider the observational data from a study that analyzes the relationship between socioeconomic factors and ecosystem services in cultural landscapes~\citep{maldonado2018}. 
In this illustrative example, only three Boolean variables are considered: \textit{Mountain ($M$)} indicating whether the topography is mountainous (yes) or flat (no); \textit{Immigration ($I$)} indicating if there are more people coming into the area (yes) than leaving it (no); and finally \textit{Agricultural-land ($A$)} indicating if the land is mainly used for agricultural activities (yes) or other activities (no). This data is summarized in Table~\ref{tab:intro_data}.

\begin{table}[htp!]
	\centering
	\caption{Data from an observational study involving three Boolean variables \citep{maldonado2018}.}
	\begin{tabular}{cccr}
		\toprule
		Mountain ($M$) & Immigration ($I$)& Agricultural-land ($A$)&Counts\\
		\midrule
		yes	&yes	&yes	&95\\
		yes	&yes	&no	&244\\
		yes	&no	&yes	&80\\
		yes	&no	&no	&183\\\hline
		no	&yes	&yes	&121\\
		no	&yes	&no	&52\\
		no	&no	&yes	&47\\
		no	&no	&no	&8\\
		\bottomrule
	\end{tabular}
	\label{tab:intro_data}
\end{table}

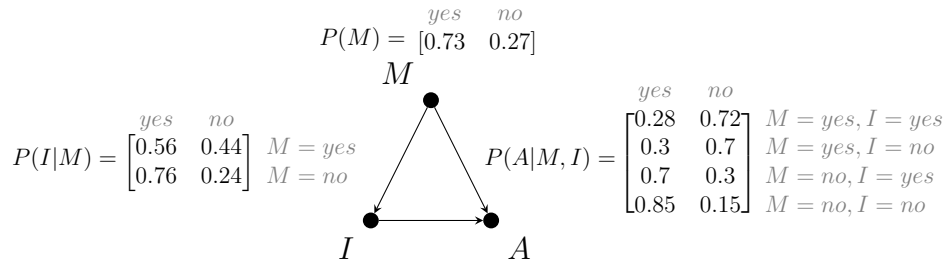
\begin{figure}[htp!]
	\centering
	\begin{tikzpicture}[scale=0.8]
		\node[dot,label=above left:{$M$}] (M)  at (1,1) {};
		\node[dot,label=below left:{$I$}] (I)  at (-0,-1) {};
		\node[dot,label=below right:{$A$}] (A)  at (2.0,-1) {};
		\draw[a2] (M) -- (I);
		\draw[a2] (M) -- (A);
		\draw[a2] (I) -- (A);

		\node[scale=0.75] at (1, 2) {$
			P(M)=
			\begin{blockarray}{cc}
				\color{gray}{yes} &  \color{gray}{no} & \\
				\begin{block}{[cc]}
					0.73 & 0.27\\
				\end{block}
			\end{blockarray}
			$};

		\node[scale=0.75] at (-3.0, -.0) {$
			P(I|M)=
			\begin{blockarray}{ccl}
				\color{gray}{yes} &  \color{gray}{no} & \\
				\begin{block}{[cc]l}
					0.56 & 0.44& {\color{gray}{M = yes} }\\
					0.76 & 0.24& \color{gray}{M = no} \\
				\end{block}
			\end{blockarray}
			$};
		
		\node[scale=0.75] at (5.8, -0) {$
			P(A|M,I)=
			\begin{blockarray}{ccl}
				\color{gray}{yes} &  \color{gray}{no} & \\
				\begin{block}{[cc]l}
					0.28 & 0.72& \tiny{\color{gray}{\tiny M = yes, I=yes} }\\
					0.3 & 0.7& \color{gray}{M = yes, I=no} \\
					0.7 & 0.3& {\color{gray}{M = no, I=yes} }\\
					0.85 & 0.15& \color{gray}{M = no, I=no} \\
				\end{block}
			\end{blockarray}
			$};
	\end{tikzpicture}
	\caption{BN obtained from the observational data in Table~\ref{tab:intro_data}.}\label{fig:into_bn}
\end{figure}

From the data presented in the table, it might be possible to build the (discrete) \emph{Bayesian network} (BN)~\citep{pearl1988} shown in Figure~\ref{fig:into_bn}. A BN is formally defined as a tuple $\langle \bmV, \mcG, \mcP_{\bmV} \rangle$ where $\bmV$ is a set of variables from the problem being modeled, $\mcG$ is a directed acyclic graph (DAG), whose nodes are the variables in $\bmV$ and $\mcP_{\bmV}$ is a set containing a conditional probability distribution $P(V | \mathrm{Pa}_V)$ for each $V\in\bmV$ where $\mathrm{Pa}_V$ are the parents of $V$ in $\mcG$.
If all the variables are discrete, the conditional distributions are represented as tables, and we will refer to them as conditional probability tables (CPTs).

Suppose our variable of interest is \textit{Agricultural-land (A)}.
When studying the impact of immigration on the land-use, one might consider to analyze the distribution $P(A|I) \propto \sum_M P(A|I,M)\cdot P(I|M) \cdot P(M)$. From the CPTs in Figure~\ref{fig:into_bn}, it follows that $P(A=yes|I=yes) = 0.42$ and $P(A=yes|I=no) = 0.39$. As there is a positive correlation between both variables, one could conclude that immigration has a positive effect on agriculture. However, analyzing separately the data in mountainous and flat areas, the correlation is the opposite as we have:  $P(A=yes|M=yes, I=yes) = 0.28 < 0.3 = P(A=yes|M=yes, I=no)$ and $P(A=yes|M=no, I=yes) = 0.7 < 0.85 = P(A=yes|M=no, I=no)$. This is an instance of the so-called Simpson's Paradox~\cite[Ch.6]{pearl2009}, which refers to a phenomenon whereby the association between a pair
of variables reverses sign upon conditioning on a third variable (a \emph{confounder}): flat areas are more likely to have immigration and also this topography is more suitable for an agricultural use of the land. A further explanation for this paradoxical situation, is that it is not the same seeing as doing. When calculating $P(A|I=yes)$, we are essentially asking about the probability of the agricultural land-use \textit{given that we see} that there is immigration. However, we might be interested in determining if the immigration is a necessary condition for the agricultural land-use. In other words, if immigration were to cease in a given area, would it lead to a reduction in agricultural land-use? Conversely, it is also valuable to understand if promoting immigration would be advantageous for agriculture, in which case we say that it is sufficient condition. These scenarios involve hypothetical situations that can be effectively addressed through counterfactual reasoning, which involves evaluating how the probability of outcomes would change if certain variables were set to specific values contrary to what actually occurred.

In line with the previous illustrative example, this paper aims to study the influence of socioeconomy on land-uses and population growth, in the conceptual framework of a socioecological system  \citep{Anderies2004}. Socioecological systems encompass the intricate interplay between human systems and natural ecosystems \citep{Berkes2003,Preise2018}. 
The socioecological system is a complex adaptive system, with  some properties, such as: non-linear dynamics, critical thresholds, tipping points, regime shifts \citep{Scheffer2012,Hughes2013,Mathias2020,ArnaizSchmitz2023},  system memory, cross-scale linkages \citep{parrott2016} and uncertainty \citep{Biggs2015}. 
All these properties are equally important for characterizing socioecological systems; however, in this work, we focus specifically on uncertainty.
In the socioecological context, land-use changes  (integrated in a landscape) are primarily
driven by socioeconomic processes, influencing the ecological integrity of these landscapes, therefore changes in socioeconomic structures and processes induce an alteration of the landscapes \citep{Schmitz2003}.

In this paper, we provide a coherent overview of the fundamental concepts for applying causal and counterfactual reasoning to data analysis within the domain of environmental and ecological sciences. In particular, we consider the use of \textit{structural causal models} (SCM)~\citep{pearl2009,bareinboim2022pearl}. SCMs are probabilistic graphical models (PGMs), i.e. they are probabilistic models in which the independence structure is encoded by a graph whose vertices are the variables in the model. SCMs are particularly designed for counterfactual reasoning and, like the rest of PGMs, they are suitable for environmental and ecological scientists and practitioners due to their graphical representation. Moreover, the recent method \textit{expectation-maximization for causal computation} (EMCC)~\citep{zaffalon2023efficient,zaffalon2023approximating} is proposed to be used for counterfactual reasoning. A key advantage of this method is its ease of implementation, primarily built upon the widely recognized \textit{expectation-maximization} (EM)~\cite[Ch.19]{koller2009probabilistic} approach for parameter learning in PGMs. We put these concepts into practical use with an observational  dataset, including information about socioeconomic factors and land-uses, in different areas of Andalusia (southern Spain). 
Unlike traditional analysis with other PGMs, the use of SCMs allows to analyze the relations of necessity and sufficiency between the variables in the aforementioned socioecological system.

This paper is structured as follows. Section~\ref{sec:related_work} reviews the relevant literature regarding the use of PGMs in the analysis of environmental data; Section~\ref{sec:background} introduces the fundamentals of causal and counterfactual reasoning, with a specific focus on SCMs and BNs; Section~\ref{sec:method} provides details about the case study considered for counterfactual analysis; the analysis of the results is presented in Section~\ref{sec:results}; finally, Section~\ref{sec:conclusions} offers an overview of the main conclusions drawn from this paper.

\section{Related work}\label{sec:related_work}

Causal modeling techniques have been used in environmental and ecological sciences in a wide range of works. \cite{paul2011} highlights the simplicity and effectiveness of causal modeling approach at dealing with confounding, in comparison to the broadly accepted before-after control-impact (BACI) designs (which are used to study ecological responses in large experimental units where replication is challenging or unfeasible \citep{green1979sampling,stewart2001temporal}). Later, \cite{paul2013} proposed ordination axes arising from multivariate macrobiotic species data in conjunction with structural equation modeling (SEM) approach to analyze the impact of the 1978 Amoco Cadiz oil spill. In this work, the conditional independencies are considered by the authors as the only means to test causal structures with observational data \citep{paul2013}. SEM are also used in \citep{paul2016} to assess the risk of wastewater discharge on macro-invertebrate communities, focusing in the adaptation of the causal diagram to a statistical model which allows for computing the effect of an intervention retrospectively. 

Structural equation modeling and BNs have proven to be useful tools to control spatiotemporal confounding in environmental studies \citep{hatami2019review,hatami2018develop,hatami2018practical}. Another work \citep{hatami2018develop} integrates BNs with SEM  to infer causal effects of wastewater on the macro-invertebrate community once the effect of natural variation is removed or to analyse the spatiotemporal variations of macrobenthic assemblage caused by leaking from a wastewater treatment plant.

Another common issue in environmental studies which is addressed by SEM is multicollinearity: \cite{bizzi2013} use SEM framework in the study of the relationships between water quality, physical habitat and benthic macroinvertebrate community in rivers, finding that SEM reduce errors due to multicollinearity thanks to the a priori selection of variables and paths. \cite{ramazi2021} use BNs to deal with missing values and highly correlated covariates in their study of a mountain pine beetle infestation. \cite{irvine2015} combine Bayesian path analysis and SEM to study the effect of stressors (such as anthropogenic drivers of road density, percent grazing or percent forest within a catchment) affect stream biological condition.

\cite{carriger2016} recommend the use of BNs for evidence-based policy in environmental management, on the grounds that these graphical models can look into the evidence for causality through improved measurements, minimizing biases in predicting or diagnosing causal relationships. In their review, the authors propose several guidance works on BN development for environmental problems and, as practical example, use BNs to study the impacts of biological and chemical stressors on a fish population.  

\cite{nyberg2006} discuss the applications of BNs to adaptive-management processes, by illustrating the system relations, calculating joint probabilities for decision options or predicting the outcomes of management policies. They exemplify the use of BNs in an adaptive management of forest and terrestrial lichens. 

Influence Diagrams (IDs) are PGMs that extend the BN by including information about the interactions of decisions with the system. Carriger et al \citep{carriger2011} develop an ID for the Deepwater Horizon spill event, which is used to display the impacts of decisions and spilled oil on ecological variables and services. This modeling framework is proposed for assistance in future responses to oil spills because it points out important variables and relationships to be accounted for by decision makers to get a better understanding of the potential outcomes. \cite{carriger2012} also propose IDs for causal inference in ecological risk-based management of pesticide usage.

Besides causal modeling, counterfactual thinking is essential in environmental policy to draw inferences about program effectiveness as well as to discriminate between program effect and biases \citep{ferraro2009}. \cite{andam2008} apply counterfactual thinking by using matching methods to improve the estimate of the impact of protected areas in Costa Rica on deforestation. In that work, the authors demonstrate that counterfactual thinking let control biases along observable features and check the sensitivity of the estimates to potential hidden biases. Also a statistical matching technique is used by \cite{mcconnachie2016} to estimate cost-effectiveness of South Africa's Working for Water program on reducing invasive species.

In relation to our case study, 
various methodologies have been employed, including multiple regression \citep{DeAranzabal2008}, econometric models \citep{Punzo2022}, graphical spatial models \citep{Irvine2011}, panel-data co-integration techniques \citep{Subramaniam2023}, and BNs \citep{aguilera2011bayesian,Ropero2019}. 
However, to the best of our knowledge, counterfactual reasoning with PGMs has not yet been explored in the context of socioecological systems. 
This approach could offer a new dimension to the study of socioeconomic influences on land-use changes within socioecological systems. In particular, in this paper we adopt the structural causal model (SCM) formalism~\citep{pearl2009}. As we will see in Section~\ref{sec:background}, SCMs can successfully handle causal reasoning from a semantic point of view. In addition, SCMs can be represented as BNs and therefore it is possible to take advantages of the existing methods for inference and learning over the latter. Another advantage is that practitioners familiarized with BNs are likely to find SCMs as a natural way of dealing with causal reasoning.

\section{Background and notation}\label{sec:background}
This section provides an overview of fundamental concepts related to PGMs for causal and counterfactual reasoning that will be later instantiated to socioecological systems. 
With respect to the general notation, upper-case letters are used to denote random variables and lower-case for their possible values (also called states), i.e. given a variable $V$, $v$ denotes an element of its domain, denoted by $\Omega_V$. We assume that all the variables are discrete. Similarly, $\bmV =\{V_1, V_2,\ldots,V_n\}$ denotes a set of variables and $\bmv$ an element of $\Omega_\bmV = \times_{V \in \bmV} \Omega_V$. The probability mass function of a discrete random variable $V$ will be denoted by $P(v)=P(V=v)$.

Causal reasoning consists of three levels~\citep{pearl2018}, namely \textit{association}, \textit{intervention} and \textit{counterfactuals}. The first level, association, accounts for predictions based on past observations. At this level, one can answer questions of the form \textit{``What if  I see ...?''}. Such questions are called \textit{observational queries} and can be answered using conditional probabilities (estimated from the observational data) stating how likely is that something happens given that something else has happened. In a general setting where $x$ and $y$ are states of the random variables $X$ and $Y$ respectively, an example of an observational query is the computation of the conditional probability $P(x|y)$. 
Herein, $ x $ and $y$ are the positive states (presence) of $X$ and $Y$, respectively, while $ x' $ and $y'$ are their counterpart negative state (absence).

The second level, intervention, is related to questions of the type \textit{``What if I do ...?''}. Such kind of questions can also be formulated in terms of probabilities using \emph{do calculus} \citep{pearl2009}. Let $Y_x$ denote the random variable representing $Y$ under the hypothetical scenario in which $X$ is forced to be equal to $x$. Then the query  $ P(Y_x=y) $ stands for the probability that $ Y $ takes the value $ y $ when $ X $ is intervened (i.e., forced) to take the value $ x $. With this in mind, we might be interested in estimating the difference between two interventional queries, which is known as the \emph{average causal effect} (ACE), defined as
\begin{equation}
	\label{eq:ace}
	\mathrm{ACE}(X,Y) = P(y_x) - P(y_{x'}) .
\end{equation}
Note that ACE takes values between -1 and 1. Positive values of ACE mean that $Y$ is more likely to happen when $X$ also happens, while negative values indicate that $Y$ is more likely to happen when $X$ does not happen.

The last level of causation, \textit{counterfactuals}, aims at queries of the form \textit{``What if I had done ...?''}. In terms of probabilities, counterfactual queries tackle hypothetical scenarios like, \textit{``What would the outcome have been if the variable had taken a different value?"}. For instance, the conditional probability $ P(Y_x | X=x') $ represents the probability of $ Y $ if $ X $ had been $ x $ instead of $ x' $.  Note that $Y_x$ is a variable related to the hypothetical scenario whereas $X$ (without subindex) denotes a variable in the real scenario. Note how counterfactual queries can give us information telling if it was $X$ that caused $Y$.

Given this semantics of interventional and counterfactual queries, it is possible to specify some typical queries that can be useful for understanding the model but also for defining policies aimed at solving problems, as we will see in the case study in Section~\ref{sec:method}. In this context, we might need to measure to what extent an event is a necessary condition for another one (i.e., when one event must occur for another one to happen). This can be achieved using the so-called  \textit{probability of necessity} (PN) which can be defined as 
\begin{equation}
    \label{eq:pn}
    \mathrm{PN}(X,Y) = P(Y_{x'}=y' | X=x,Y=y) .
\end{equation}

$X$ is said to be a necessary cause for $Y$ if whenever $y$ occurs then $x$ has occurred. Therefore, PN can be interpreted as the probability that $X$ is a necessary cause of $Y$.
In other words, it is the probability that the event $y$ would not have occurred in the absence of event $x$, given that $x$ and $y$ did in fact occur.
In our running example, $\mathrm{PN}(I, A)$ measures the degree of certainty with which we can assert that whenever agriculture is present then immigration is also present.

It might also be useful to consider the probability of necessity, but assuming that $X=x$ did not happen. We call it the \emph{probability of necessity with reverse cause} (PNrc), formally defined as
	\begin{equation}
		\label{eq:pnrc}
		\mathrm{PNrc}(X,Y) = P(Y_{x}=y' | X=x',Y=y) .
	\end{equation}

Analogously, we could also be interested in determining if an event is a sufficient condition for another event to happen. For this we can define the \textit{probability of sufficiency} (PS) as

\begin{equation}
    \label{eq:ps}
    \mathrm{PS}(X,Y) = P(Y_{x}=y | X=x',Y=y') .
\end{equation}

$X$ is said to be a sufficient cause for $Y$ if whenever $x$ occurs then $y$ will occur. Therefore, PS can be interpreted as the probability that $X$ is a sufficient cause of $Y$. 
In other words, it is the probability that setting $x$ would produce $y$ in a scenario where $x$
and $y$ are in fact absent. 
In our example, $\mathrm{PS}(I, A)$ measures the degree of certainty with which we can assert that whenever immigration is present, agriculture is also present.

An event could also be, to some extent, both necessary and sufficient. It can be measured by the \textit{probability of necessity and sufficiency} (PNS), defined as
	\begin{equation}
		\label{eq:pns}
		\mathrm{PNS}(X,Y) = P(Y_{x}=y , Y_{x'}=y') .
	\end{equation}

$X$ is said to be a necessary and sufficient cause for $Y$ if whenever $x$ occurs then $y$ will occur and vice-versa. Therefore, PNS can be interpreted as the probability that $X$ is a necessary and sufficient cause of $Y$.  Intuitively, PNS measures how $Y$ reacts to $X$, hence expressing  to what extent $X=x$ is necessary and sufficient for $Y=y$. In the example provided, $\text{PNS}(I,A)$ measures the degree of certainty with which we can assert that whenever immigration is present, agriculture is also present, and whenever immigration is absent, agriculture is also absent.

The first level of causation (association) can be properly handled using probability distributions represented as a Bayesian network, where all the possible observational queries can be answered by computing the relevant conditional probabilities directly on the network. However, handling the other two levels of causation (interventions and counterfactuals) requires going beyond conditional probabilities, so that we are able to handle hypothetical (and not only observed) scenarios. In order to approach these kind of scenarios, \cite{pearl2009} proposed the so-called structural causal models (SCMs), which are a specific type of probabilistic graphical model used for causal and counterfactual reasoning. SCMs distinguishes between two types of nodes: \emph{endogenous} nodes, which represent the variables within the modeled problem, for which data is available, and \emph{exogenous} nodes, which are associated with external factors for which data is not available. Note that the terms node and variable are used interchangeably. It can be shown that SCMs can be regarded as Bayesian networks that have been extended to accommodate the exogenous variables (see~\ref{ap:scm} for the technical details). 

An example of how counterfactual queries are computed over SCMs is given in~\ref{ap:queries}.
Note that if all the conditional probability distributions (CPTs) involved in an SCM are known, all the counterfactual queries described above can be answered by using standard inference algorithms for BNs. However, in problems where only observational data is available, and particularly in socioecological systems, the parameters of the CPTs corresponding to the exogenous variables cannot be uniquely determined, because there is no data about those variables. When this problem arises, it is said that the counterfactual query is \emph{unidentifiable}~\citep{correa2021nested,wu2019counterfactual}.

Nonetheless there are methods able to deal with problems where only observational data is available, but instead of precise probability values, they provide intervals bounding them~\citep{tian2000probabilities,zaffalon2020structural}. In our specific case study,  we propose the utilization of the innovative technique known as EMCC (\textit{Expectation Maximization for Causal Computation}) as detailed by \cite{zaffalon2023approximating,zaffalon2023efficient}.

\section{Case study}\label{sec:method}
\subsection{Problem and data description}
\begin{figure}[h]
	\centering
	\includegraphics[width = 0.8\textwidth]{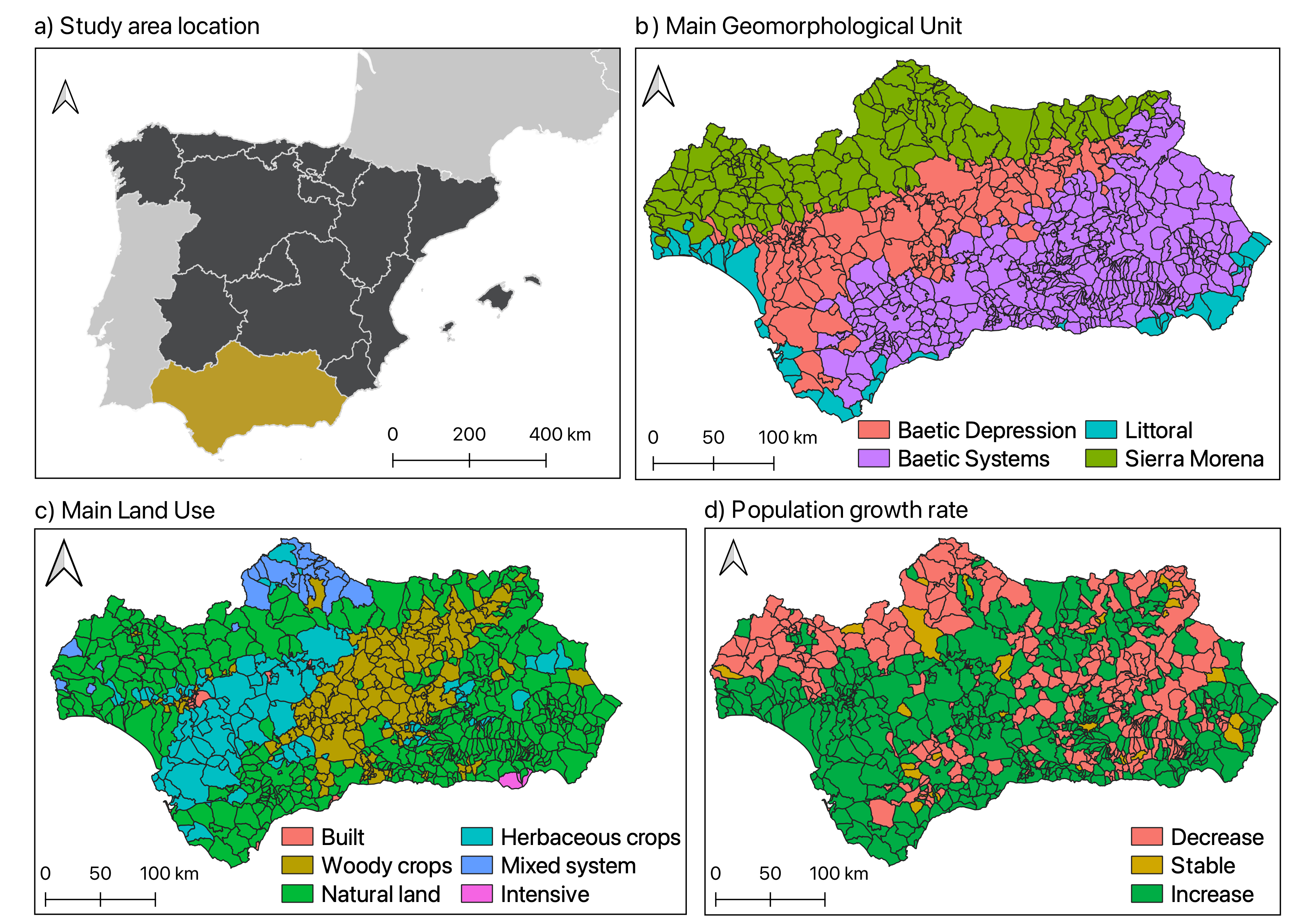}
	\caption{Study area (Andalusia, Spain) (a) and municipalities within the study area, color-coded based on their primary geomorphological unit (b), their main land-use (c), and their population growth rate (d). For (b) and (c), in cases where a municipality encompasses more than one geomorphological unit or land-use, the color represents the larger or dominant one within that municipality.}
	\label{fig:maps}
\end{figure}

To illustrate the potential of counterfactual reasoning with PGMs, let us consider the ecological and socio-economic data chosen for the case study, which is related to the region of Andalusia, in southern Spain (Figure \ref{fig:maps}~(a)). This area of study is a socioecological system with a strong relation between the natural and socio-economic components.
It also shows high variability regarding elevation, ranging from 0 to 3460 m above the sea level. The main mountain ranges within the study area are the Sierra Morena mountain range in the North and the Baetic Systems in the South, with the Baetic Depression serving as the geological boundary between them. The Guadalquivir River flows through this depression, being the largest river in Andalusia. The flattest areas correspond to the littoral and the Baetic depression, while the steepest ones correspond to the Baetic Systems. Therefore, Andalusia can be divided into 4 main geomorphological units: the Baetic Depression, the Sierra Morena mountain range, the Baetic Systems and the Littoral, as shown in Figure~\ref{fig:maps}~(b).

The Baetic Depression is characterized by its high agricultural production, mainly comprising rain-fed herbaceous crops in the low-lying plains and irrigated herbaceous crops along the Guadalquivir river (Figure~\ref{fig:maps}~(c)). The Sierra Morena mountain range is characterized by having high emigration and mortality rates and low birth rate, which results in population decline (Figure~\ref{fig:maps}~(d)), and is predominantly covered by rain-fed crops and dehesa, a heterogeneous system exhibiting various states of ecological maturity, with shepherding being the principal economic activity. The Baetic Systems have the highest elevation and steepness in the study area. This area is predominantly cloaked in natural vegetation, with extensive woody crops as a secondary feature. Its rugged terrain discourages the adoption of intensive agricultural practices. Finally, the Littoral, densely populated and characterized by high temperatures, features abundant natural vegetation and serves as the primary location for the majority of greenhouses in the study area, making it a focal point of agricultural activity.

In a previous study~\citep{maldonado2018}, 75 different variables representing social, economic and ecological characteristics of the study area were employed to study how socioeconomic changes influence the generation of ecosystem services
using BNs with no causal (nor counterfactual) reasoning conducted at that time. 
These variables, available in public repositories, were sourced from the Multi-territorial Information System of Andalusia (SIMA) and the Andalusian Environmental Information Network, and municipalities within the study area were taken as the modeling unit. Conversely, in our current study, we narrowed our focus to a subset of 17 variables from the original dataset, to conduct a causal and counterfactual analysis using SCMs. These include land-use, social, and economic variables which are detailed in Tables \ref{tab:ecological}, \ref{tab:social}, and \ref{tab:economic}, respectively.

\begin{small}
	\footnotesize
	\renewcommand{\arraystretch}{1.8}
	\linespread{0.9}\selectfont\centering
	
	\begin{longtable}{ N   L{0.5in}  L{2.4in} L{0.8in} L{0.7in}  N}
		\caption{Variables representing the ecological dimension used to construct the SCM. }
		\label{tab:ecological}\\
		\hline
		& \multicolumn{4}{c}{\bf Ecological dimension}   & \\ \hline
		&  {\bf Name} & {\bf Description} &{\bf State}&{\bf Threshold}& \\\hline
		\endhead 
		& MGU & The main geomorphological unit a municipality belongs to.&Baetic~Depr. \newline Sierra~Morena \newline Baetic Sys. \newline Littoral&& \\
		& Built& Percentage of artificial or built areas in each municipality - including {\em urban; industrial; mining; freight and technical infrastructures}.&Scarce \newline Fair \newline Abundant&$<$ 5 \newline 5 - 30 \newline $>$ 30& \\
		& GH& Percentage of intensive agriculture (greenhouses) in each municipality.&Scarce \newline Fair \newline Abundant&$<$ 5 \newline 5 - 30 \newline $>$ 30& \\
		& HCrops&Percentage of herbaceous crops in each municipality -  including rainfed and irrigated crops.&Scarce \newline Fair \newline Dominant&$<$ 15 \newline 15 - 50 \newline $>$ 50& \\
		& WCrops& Percentage of woody crops in each municipality -  including rainfed and irrigated crops.&Scarce \newline Fair \newline Dominant&$<$ 15 \newline 15 - 50 \newline $>$ 50& \\
		& Mixed  & Percentage of heterogeneous lands in each municipality -  including patches mixing {\em grassland and forest} and {\em crops with natural vegetation}.&Scarce \newline Fair \newline Dominant&$<$ 10 \newline 10 - 40 \newline $>$ 40& \\
		& Natural& Percentage of natural areas in each municipality - comprising {\em bush; grassland; forest; bush and forest; wetlands and naked soil}.&Scarce \newline Fair \newline Dominant&$<$ 25 \newline 25 - 60 \newline $>$ 60& \\\hline
	\end{longtable}
	\begin{longtable}{ N   L{0.5in}  L{2.5in} L{0.6in} L{0.7in} N}
		\caption{Variables representing the social dimension used to construct the SCM. } \label{tab:social} \\
		\hline
		& \multicolumn{4}{c}{\bf Social dimension}& \\ \hline
		&  {\bf Name} & {\bf Description} &{\bf State}&{\bf Threshold}& \\\hline
		\endhead 
		
		& Pop& Population density of each municipality in 2011 (inhabitants/$Km^2$).&Low \newline Moderate \newline High&$<$ 35 \newline 35- 150 \newline $>$ 150& \\
		& SR& Sex ratio. Proportion of males (M) to females (F) in each municipality in 2011 (computed as $SR = \frac{M}{M+F}$).&More females \newline More males&$\leq$ 0.50 \newline $>$ 0.50&\\
		& EGR& Population growth rate. Exponential growth of the population, computed as  $EGR=\frac{\ln (P_t / P_0)}{t}$, where $P_0$ represents the population in 2001, $P_t$ the population in 2011 and $t$ the 10-year period.&Decrease \newline Stable \newline Increase&$<$ -0.03 \newline -0.03 - 0.03 \newline $>$ 0.03&\\
		& IME& Index of Migration effectiveness. Percentage of total migration for the period 2001-2011. It ranges from -100 (emigration) to 100 (immigration),  with values close to 0 indicating no change in the population dynamic. It is computed as $IME = \frac{Immigration-Emigration}{Immigration+Emigration}\times100$.  &Emigration \newline Balanced \newline Immigration&$<$ -2 \newline -2 - 2 \newline $>$ 2&\\
		& ODI& Old-age dependency index. Percentage of the older over the younger population in 2011,  computed as $ODI = \frac{P_{>65}}{P_{<15}}\times100$, where $P_{>65}$ is the population older than 65 years old and $P_{<15}$ is the population younger than 15 years old.&Low \newline Moderate \newline High&$<$ 25 \newline 25 - 40 \newline $>$ 40&\\
		& Death  & Mortality rate. Number of deaths per 1000 inhabitants in each municipality in 2011.&Low \newline Moderate \newline High&$<$ 9 \newline 9 - 15 \newline $>$ 15& \\
		& Birth& Birth rate. Number of births per 1000 inhabitants in each municipality in 2011. &Low \newline Moderate \newline High&$<$ 5.6 \newline 5.6 - 10.3 \newline $>$ 10.3& \\
		\hline
	\end{longtable}
	\begin{longtable}{ N  L{0.5in}  L{2.5in} L{0.6in} L{0.7in} N}
		\caption{Variables representing the economic dimension used to construct the SCM. }\label{tab:economic}\\
		\hline
		& \multicolumn{4}{c}{\bf Economic dimension}& \\ \hline
		&  {\bf Name} & {\bf Description} &{\bf State}&{\bf Threshold}& \\\hline
		\endhead 
		&WF&Workforce. Percentage of the municipality's working age population ($\geq 16$) that are available to work in 2011. It is computed as $WF= \frac{ER + UR}{P_{\geq16}}\times 100$;  where $ER$ is the Employment Rate;  $UR$ is the Unemployment Rate and $P_{\geq16}$ is the population older than 16 years old.&Low \newline Average \newline High&$<$ 0.55 \newline 0.55 - 0.62 \newline $>$ 0.62& \\
		& SSE & Secondary sector employment. Number of employees in the secondary sector per 1000 inhabitants.  &Low \newline Moderate \newline High&$<$ 70 \newline 70 - 113 \newline $>$ 113& \\
		& TSE  &Tertiary sector employment. Number of employees in the trading, banking and service sectors per 1000 inhabitants. &Low \newline Moderate \newline High&$<$ 78 \newline 78 - 122 \newline $>$ 122& \\
		\hline
	\end{longtable}
\end{small}

The primary goal of this study is to investigate how different variables of interest in this socioecological system are influenced by other variables. In this context, such variables of interest are termed \textit{effect variables}, while the other factors that could potentially influence these effects are referred to as \textit{cause variables}. In connection to Section \ref{sec:background}, $X$ and $Y$ represent the cause and effect variables, respectively. Specifically, we consider as effects those variables representing the different land-uses and the population growth (i.e., \textit{Built}, \textit{GH}, \textit{HCrops}, \textit{WCrops}, \textit{Mixed}, \textit{Natural} and \textit{EGR}). We will refer to the union of the cause and effect variable sets as the \textit{variables of interest}.

\subsection{Data preprocessing and model definition}

To conduct any causal or counterfactual analysis, it is necessary to define the cause and effect variables in such a way that their values can be categorized into two groups. In this case, since the variables are categorical (with multiple states), they are transformed into a binary format by collapsing their values into two distinct states: one representing a positive outcome and the other a negative outcome. Table~\ref{tab:binarization} shows the partitions of the states considered for each variable in the dataset.

\begin{table}[h]
	\centering
	\caption{Categorization into positive and negative values.}
	\label{tab:binarization}
	\begin{tabular}{lll}
		\hline
		Variable  & Positive state          & Negative state                   \\\hline
		MGU       & Littoral, Baetic Depression & Baetic System, Sierra Morena \\
		Built     & Abundant, Fair          & Scarce                           \\
		GH        & Abundant, Fair          & Scarce                           \\
		Hcrops    & Dominant, Fair          & Scarce                           \\
		Wcrops    & Dominant, Fair          & Scarce                           \\
		Mixed     & Dominant, Fair          & Scarce                           \\
		Natural   & Dominant, Fair          & Scarce                           \\
		Pop     & High                    & Low, Moderate                    \\
		SR        & More females            & More males                       \\
		EGR       & Increase                & Decrease, Stable                 \\
		IME       & Immigration             & Emigration, Balanced             \\
		ODI       & Low                     & Moderate, High                   \\
		Death & Low                     & Moderate, High                   \\
		Birth     & High                    & Low, Moderate                    \\
		WF        & High                    & Low, Average                     \\
		SSE       & High                    & Low, Moderate                    \\
		TSE       & High                    & Low, Moderate                   \\\hline
	\end{tabular}
\end{table}

Given that the counterfactual analysis relies on SCMs, a causal structure in the form of a DAG is required. To establish this structure, we adopt the BN structure used in the aforementioned prior investigation~\citep{maldonado2018}, initially formulated by domain experts. 
In that work, the approach used to build the BN was based on the DPSIR framework \citep{EEA_4assess07}. In this context, socioeconomic variables are considered the Drivers (as defined in the DPSIR framework) of environmental change. These Drivers produce different Pressures, which are reflected as changes in land use, ultimately altering the State of the ecosystem. This, in turn, affects the supply of ecosystem services and human well-being (Impacts). Finally, Responses are the different actions that governments and society take to control the Drivers. To align with this framework, meaningful variables and their causal relationships were identified through expert knowledge and relevant literature.

\begin{figure}[h]
	\centering
	\resizebox{0.85\textwidth}{!}{
		\begin{tikzpicture}[scale=0.7]
			\node[dot,label=above:{$MGU$}] (MGU)  at (0.5,2) {};
			\node[dot,label=above:{$SSE$}] (SSE)  at (-6,0) {};
			\node[dot,label=above:{$TSE$}] (TSE)  at (-2.5,-0) {};
			\node[dot,label= left:{$IME$}] (IME)  at (-8,-2.5) {};
			\node[dot,label=above right:{$\;\;\;\;ODI$}] (ODI)  at (-4,-2.5) {};
			\node[dot,label=right:{$SR$}] (SR)  at (-1,-2.5) {};
			\node[dot,label=below left:{$WF$}] (WF)  at (-8,-4) {};
			\node[dot,label=right :{$Death$}] (Death)  at (-5.5,-4) {};
			\node[dot,label=right:{$Birth$}] (Birth)  at (-2.5,-4) {};
			\node[dot,label=below :{$EGR$}] (EGR)  at (-4,-5.5) {};
			\node[dot,label=below :{$Pop$}] (PDens)  at (0.5,-7) {};
			\node[dot,label=below :{$Built$}] (Built)  at (1.5,-2.0) {};
			\node[dot,label=below :{$GH$}] (GH)  at (2.7,-2.0) {};
			\node[dot,label=below :{$HCrops$}] (HCrops)  at (4.2,-2.0) {};
			\node[dot,label=below  :{$Mixed$}] (Mixed)  at (6,-2.0) {};
			\node[dot,label=below :{$Natural$}] (Natural)  at (8,-2.0) {};
			\node[dot,label=below  :{$WCrops$}] (WCrops)  at (10,-2.0) {};
			\draw[a2] (MGU) -- (SSE);
			\draw[a2] (MGU) -- (TSE);
			\draw[a2] (MGU) to [bend right=45] (IME);
			\draw[a2] (MGU) -- (SR);
			\draw[a2] (MGU)  -- (ODI);
			\draw[a2] (TSE) -- (SR);
			\draw[a2] (SSE) -- (ODI);
			\draw[a2] (TSE) -- (ODI);
			\draw[a2] (MGU) to [bend right=-3] (Built);
			\draw [a2] (MGU) to [bend right=-8] (GH);
			\draw [a2] (MGU) to [bend right=-15] (HCrops);
			\draw [a2] (MGU) to [bend right=-25] (Mixed);
			\draw [a2] (MGU) to [bend right=-25] (Natural);
			\draw [a2] (MGU) to [bend right=-35] (WCrops);
			\draw[a2] (SSE) -- (IME);
			\draw[a2] (TSE) -- (IME);
			\draw[a2] (IME) -- (WF);
			\draw[a2] (ODI) -- (WF);
			\draw[a2] (ODI) -- (Death);
			\draw[a2] (ODI) -- (Birth);
			\draw [a2] (ODI) to [bend right=-45] (PDens);
			
			\draw[a2] (SR) -- (Birth);
			\draw[a2] (IME) to [bend right=10]  (EGR);
			\draw[a2] (Death) -- (EGR);
			\draw[a2] (Birth) -- (EGR);
			\draw[a2] (WF) to [bend right=25] (PDens);
			\draw [a2] (EGR) -- (PDens);
			\draw[a2] (PDens) to [bend right=3] (Built);
			\draw [a2] (PDens) to [bend right=8] (GH);
			\draw [a2] (PDens) to [bend right=15] (HCrops);
			\draw [a2] (PDens) to [bend right=25] (Mixed);
			\draw [a2] (PDens) to [bend right=25] (Natural);
			\draw [a2] (PDens) to [bend right=35] (WCrops);
			
		\end{tikzpicture}
	}
	\caption{BN obtained from the model in a previous study~\citep{maldonado2018} by restricting it to our variables of interest and by limiting the number of parents to 3.}
	\label{fig:exp_bn}
\end{figure}
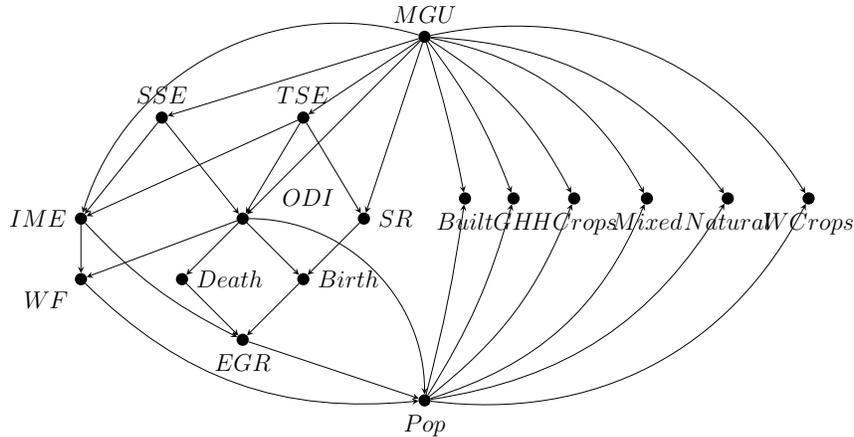

Nonetheless, in this new work some modifications are made on the structure of the original BN with the purpose of reducing its complexity, resulting in the graph shown in Figure~\ref{fig:exp_bn}. 
First, the graph is restricted to our effect variables and their ancestors, i.e. the variables of interest previously detailed in Tables \ref{tab:ecological}, \ref{tab:social}, and \ref{tab:economic}. 
Note that any descendant from the effect variables in the original graph is irrelevant for a causal analysis. Additionally, the node \textit{MGU} is a common ancestor of all the cause and effect variables considered, and there exists no alternative path connecting them with the ancestors excluded from the final graph. Consequently, our variables of interest are independent from the rest of ancestors given \textit{MGU}. Secondly, the number of parents is limited to a maximum of three. The intention of this is to prevent extremely large exogenous variables, whose number of states increases exponentially with the number of parents. \ref{sec:complexity} provides a further discussion about this. The choice of a maximum of three parents was made based on preliminary experiments, which showed that the computation with larger models was extremely time-consuming. Hence, the arc from \textit{MGU} to \textit{Pop} was removed. Note that there are alternative directed paths between both nodes, i.e., it is redundant. Moreover, this deletion is the one involving the lowest decrease in the likelihood of the model given the data.

\begin{figure}[h!tb]
	\centering
	\resizebox{0.85\textwidth}{!}{
		\begin{tikzpicture}[scale=0.7]
			\node[dot,label=above left:{$MGU$}] (MGU)  at (0.5,2) {};
			\node[dot,label=above:{$SSE$}] (SSE)  at (-6,0) {};
			\node[dot,label=above:{$TSE$}] (TSE)  at (-2.5,-0) {};
			\node[dot,label= left:{$IME$}] (IME)  at (-8,-2.5) {};
			\node[dot,label=above right:{$\;\;\;\;ODI$}] (ODI)  at (-4,-2.5) {};
			\node[dot,label=right:{$SR$}] (SR)  at (-1,-2.5) {};
			\node[dot,label=below:{$WF$}] (WF)  at (-8,-4) {};
			\node[dot,label=right :{$Death$}] (Death)  at (-5.5,-4) {};
			\node[dot,label=right:{$Birth$}] (Birth)  at (-2.5,-4) {};
			\node[dot,label=below :{$EGR$}] (EGR)  at (-4,-5.5) {};
			\node[dot,label=below :{$Pop$}] (PDens)  at (0.5,-7) {};
			
			\node[dot,label=below :{$Built$}] (Built)  at (1.5,-2.0) {};
			\node[dot,label=below :{$GH$}] (GH)  at (2.7,-2.0) {};
			\node[dot,label=below :{$HCrops$}] (HCrops)  at (4.2,-2.0) {};
			\node[dot,label=below  :{$Mixed$}] (Mixed)  at (6,-2.0) {};
			\node[dot,label=below :{$Natural$}] (Natural)  at (8,-2.0) {};
			\node[dot,label=below  :{$WCrops$}] (WCrops)  at (10,-2.0) {};
			
			\node[dot2, above right = 20pt of MGU, label=above:{$H_{MGU}$}] (HMGU) {};
			\node[dot2, above right = 20pt of SSE, label=above right:{$H_{SSE}$}](HSSE){};
			\node[dot2, above = 22pt of TSE, label=above:{$H_{TSE}$}] (HTSE){};
			\node[dot2,above left = 20pt of IME,label= left:{$H_{IME}$}] (HIME)  {};
			\node[dot2,left = 20pt of ODI,label=above:{$H_{ODI}$}] (HODI) {};
			\node[dot2, above  = 20pt of SR, label=above:{$H_{SR}$}] (HSR){};
			\node[dot2,left = 20pt of WF, label=below left:{$H_{WF}$}] (HWF){};
			\node[dot2, below left = 35pt of Death, label=below :{$H_{Death}$}] (HDeath){};
			\node[dot2, below right = 20pt of Birth, label=right:{$H_{Birth}$}] (HBirth) {};
			\node[dot2,right = 15pt of EGR,label=right :{$H_{EGR}$}] (HEGR) {};
			\node[dot2, above left = 10pt and 12pt of PDens, label=above :{$H_{Pop}$}] (HPDens) {};
			
			\node[dot2, above right = 20pt and 1pt of Built, label=above:{$H_{Built}$}] (HBuilt) {};
			\node[dot2, above right = 20pt and 2pt of GH, label=above:{$H_{GH}$}] (HGH) {};
			\node[dot2, above right = 20pt and 2pt of HCrops, label=above:{$H_{HCrops}$}] (HHCrops) {};
			\node[dot2, above right = 20pt and 2pt of Mixed, label=above:{$H_{Mixed}$}] (HMixed) {};
			\node[dot2, above right = 20pt and 2pt of Natural, label=above:{$H_{Natural}$}] (HNatural) {};
			\node[dot2, above right = 20pt and 2pt of WCrops, label=above:{$H_{WCrops}$}] (HWCrops) {};
			\draw[a2] (MGU) -- (SSE);
			\draw[a2] (MGU) -- (TSE);
			\draw[a2] (MGU) to [bend right=45] (IME);
			\draw[a2] (MGU) -- (SR);
			\draw[a2] (MGU)  -- (ODI);
			\draw[a2] (TSE) -- (SR);
			\draw[a2] (SSE) -- (ODI);
			\draw[a2] (TSE) -- (ODI);
			\draw[a2] (MGU) to [bend right=-3] (Built);
			\draw [a2] (MGU) to [bend right=-8] (GH);
			\draw [a2] (MGU) to [bend right=-15] (HCrops);
			\draw [a2] (MGU) to [bend right=-25] (Mixed);
			\draw [a2] (MGU) to [bend right=-25] (Natural);
			\draw [a2] (MGU) to [bend right=-35] (WCrops);
			\draw[a2] (SSE) -- (IME);
			\draw[a2] (TSE) -- (IME);
			\draw[a2] (IME) -- (WF);
			\draw[a2] (ODI) -- (WF);
			\draw[a2] (ODI) -- (Death);
			\draw[a2] (ODI) -- (Birth);
			\draw [a2] (ODI) to [bend right=-45] (PDens);
			\draw[a2] (SR) -- (Birth);
			\draw[a2] (IME) to [bend right=10]  (EGR);
			\draw[a2] (Death) -- (EGR);
			\draw[a2] (Birth) -- (EGR);
			\draw[a2] (WF) to [bend right=25] (PDens);
			\draw [a2] (EGR) -- (PDens);
			\draw[a2] (PDens) to [bend right=3] (Built);
			\draw [a2] (PDens) to [bend right=8] (GH);
			\draw [a2] (PDens) to [bend right=15] (HCrops);
			\draw [a2] (PDens) to [bend right=25] (Mixed);
			\draw [a2] (PDens) to [bend right=25] (Natural);
			\draw [a2] (PDens) to [bend right=35] (WCrops);
			\draw[a] (HMGU) -- (MGU);
			\draw[a] (HSSE) -- (SSE);
			\draw[a] (HTSE) -- (TSE);
			\draw[a] (HIME) -- (IME);
			\draw[a] (HODI) -- (ODI);
			\draw[a] (HSR) -- (SR);
			\draw[a] (HWF) -- (WF);
			\draw[a] (HDeath) -- (Death);
			\draw[a] (HBirth) -- (Birth);
			\draw[a] (HEGR) -- (EGR);
			\draw[a] (HPDens) -- (PDens);
			\draw[a] (HBuilt) to [bend right=3] (Built);
			\draw [a] (HGH) to [bend right=0] (GH);
			\draw [a] (HHCrops) to [bend right=0] (HCrops);
			\draw [a] (HMixed) to [bend right=0] (Mixed);
			\draw [a] (HNatural) to [bend right=0] (Natural);
			\draw [a] (HWCrops) to [bend right=0] (WCrops);
		\end{tikzpicture}
	}
	\caption{Markovian SCM used for intended counterfactual analysis. All the SEs are asumed to be canonical. Each endogenous variable has only one exogenous cause, and each exogenous variable is cause of only one endogenous one.}
	\label{fig:exp_scm}
\end{figure}
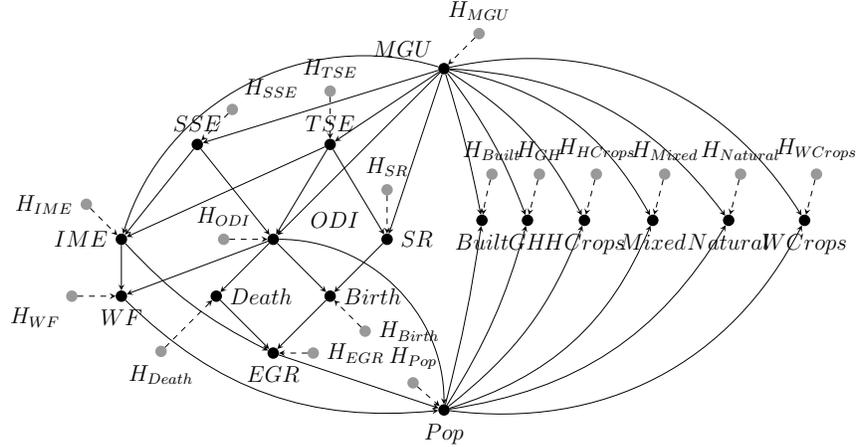

After establishing the causal structure, a SCM can be derived by introducing an exogenous parent for each endogenous variable. The resulting graph is illustrated in Figure \ref{fig:exp_scm}. As for the parameters, the SEs are defined in a canonical form. By contrast, the marginal distributions over the exogenous variables are considered to be unknown and estimated using EMCC. Specifically, this method is executed with 100 EM runs while capping the EM convergence at $300$ iterations. 

Initially, we examine the difference in conditional probability, denoted as $P(y|x) - P(y|x')$. This aligns with the conventional BN analysis. Moving to causal (non-counterfactual) reasoning, we investigate the interventional query ACE, defined in Eq.~\eqref{eq:ace}. To underscore the advantages of counterfactual reasoning, we evaluate the queries PN, PNrc, PS, and PNS, outlined in \crefrange{eq:pn}{eq:pns}. The variables under study (i.e., effects) are those enumerated in the preceding section, while the causes vary based on the specific effect. This distinction arises from the requirement that causes must be ancestors of the effects. In this sense, in the case of land-use variables, the causes include $WF$, $EGR$, $MGU$, $SR$, $SSE$, $TSE$, $ODI$, $Pop$, $Death$, and $Birth$. When $EGR$ is treated as the effect, $Pop$ and the variable itself are excluded as potential causes.

In relation to the implementation, the CREDICI software \citep{cabanas2020credici} was used, a Java library designed for causal reasoning, featuring the implementation of EMCC. For more extensive implementation details and the code to replicate the study, please refer to our GitHub repository\footnote{\url{https://github.com/PGM-Lab/2024-counterfactual-land}}. The computation was made in a computer cluster made of 1024 cores (2048 threads). This allowed the parallelization of the execution of each of the 100 EM runs, each of them taking more than 4 hours on average.

\section{Results and discussion}\label{sec:results}

Before we present the experimental results, we give a brief explanation of how to interpret them. As indicated at the end of Section~\ref{sec:background}, the results are given as probability intervals rather than as single values. Note that, a variable is considered to be a necessary or sufficient cause (or both) of another if the associated probability interval is high and narrow. For instance, a probability of necessity in the interval [0, 0.99] is too wide to be informative. On the other hand, an interval of [0, 0.1] indicates with high certainty that the variable is not necessary for the given effect. Similarly, an interval of [0.9, 0.99] suggests strong evidence that the variable is necessary for the effect to occur.

\begin{figure}[hbt]
	\centering
	\includegraphics[width = .99\textwidth]{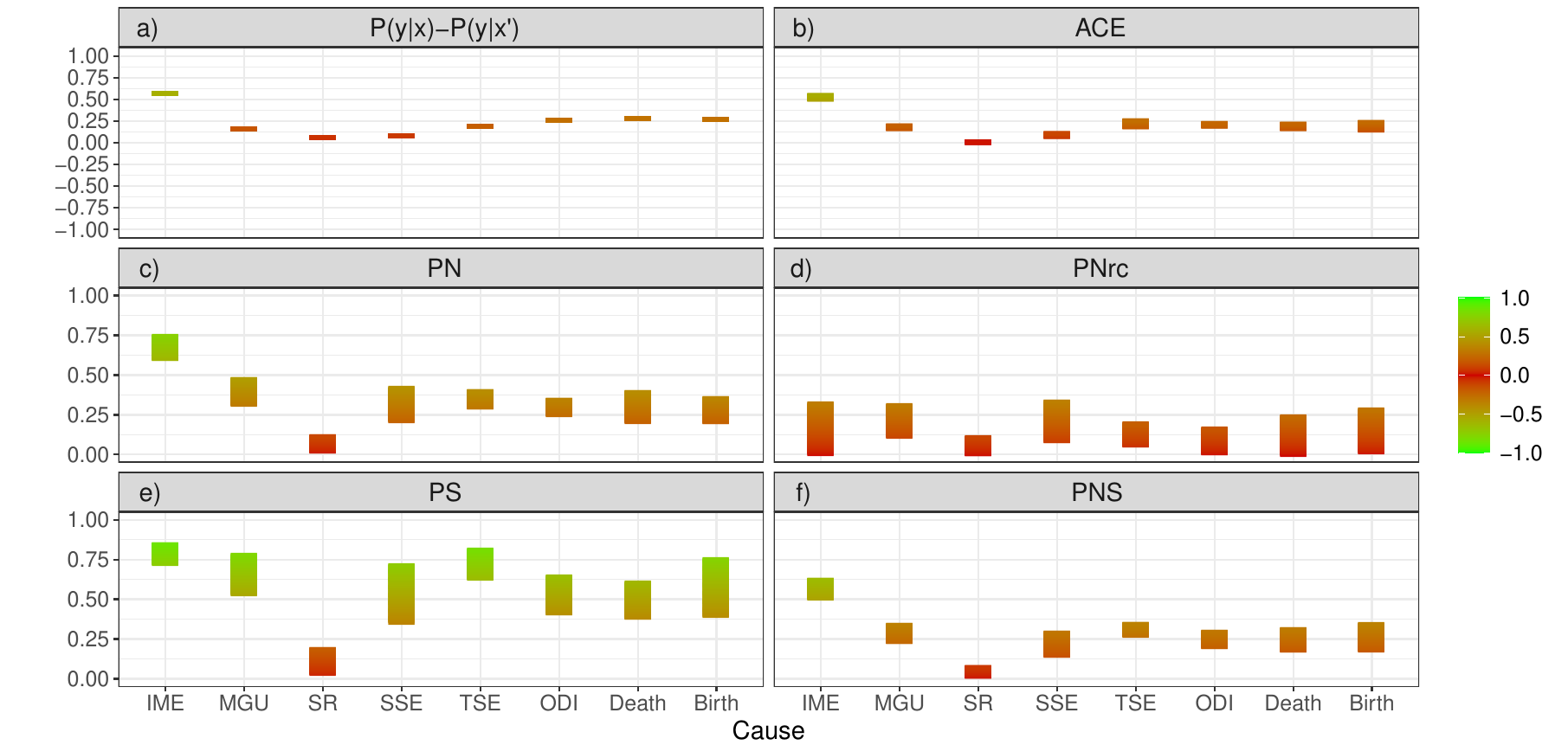}
    \caption{Intervals for the queries with \textit{EGR} as effect variable. The x-axis represents each cause variable, according to the graph in Figure~\ref{fig:exp_bn}, while the y-axis shows the query metric. Panels (a) to (f) depict different types of analysis: (a) conventional BN analysis, (b) causal analysis, and (c-f) counterfactual analysis. The metrics illustrated are (a) the difference in conditional probability, $P(y|x)-P(y|x')$; (b), the average causal effect, ACE; (c) the probability of necessity, PN; (d) the probability of necessity with reverse cause, PNrc; (e) the probability of sufficiency, PS; and (f) the probability of necessity and sufficiency, PNS. Note that metrics in panels (a) and (b) can take negative values, as they are defined as differences of probabilities.}
	\label{fig:EGR}
\end{figure}

To begin with the analysis of the experimental results, we first consider as \textit{effect} the variable \textit{EGR} (population growth rate), whose results are given in Figure~\ref{fig:EGR}. The variable with a clearer causal impact on \textit{EGR} is \textit{IME} (index of migration effectiveness). 
Remarkably, the probability of sufficiency, PS, is bounded between $0.73$ and $0.85$ (Figure~\ref{fig:EGR}~(e)), which means that it is very probable that a positive value of \textit{IME} (immigration) is enough to produce a positive value of \textit{EGR} (increase) despite the values of the other variables. Besides, it is also quite likely that \textit{IME} is a necessary condition for \textit{EGR} to have a high value, since $\mathrm{PN}(IME,EGR)\in [0.60,0.75]$ (Figure~\ref{fig:EGR}~(c)). The probability of necessity and sufficiency is also bounded above $0.5$, more precisely in the interval $[0.51,0.63]$ (Figure~\ref{fig:EGR}~(f)). 

Regarding the probability of sufficiency of the other variables under consideration, \textit{MGU} (main geomorphological unit) and \textit{TSE} (tertiary sector employment) are likely to be sufficient for \textit{EGR} to take place, reaching values in the intervals $[0.54,0.78]$ and $[0.63,0.82]$, respectively. Conversely, \textit{SSE} (secondary sector employment), \textit{ODI} (old-age dependency index), \textit{Death} (death rate) and \textit{Birth} (birth rate) have a good chance to be sufficient, since their lower bounds are close to $0.4$ (Figure~\ref{fig:EGR}~(e)).

With respect to non-counterfactual queries (Figure~\ref{fig:EGR}~(a-b)), the difference in conditional probability ($0.58$) and the average causal effect in the interval $[0.50,0.56]$ support the classification of positive \textit{IME} as a cause of positive \textit{EGR}. Note that while non-counterfactual queries enable the identification of variables influencing \textit{EGR}, these queries do not provide information about the nature of the relationship (whether it is one of necessity or sufficiency). This underscores the nuanced insights that counterfactual reasoning can offer in understanding causal relationships within the studied context.

Therefore, a positive immigration rate turns out to be both necessary and sufficient (with high probability) in order to achieve a positive population growth rate \citep{parsons2006,vinuela2019,vinuela2022}. 
On the other hand, it is also likely that the location of the municipality and the tertiary sector employment can cause a positive \textit{EGR} value, but this is only expressed in terms of sufficiency. 
As a matter of fact, Figure~\ref{fig:maps}~(d) indicates a population decline in the majority of municipalities situated in both the Baetic Systems and Sierra Morena regions.
Considering the remaining causes, population growth can occur in the absence of all of them, though they might be sufficient on their own to drive population growth. Population growth is a phenomenon influenced by various factors, including immigration, emigration, death, and birth rates \citep{lutz2006,poston2010}. The variable \textit{IME} is an index that incorporates both immigration and emigration rates. A high value of \textit{IME} indicates more immigration than emigration, resulting in population increase, assuming that other factors remain constant. In contrast, death and birth rates are considered separately rather than jointly in a single rate of natural population increase. Individually, neither low death rate nor high birth rate is necessary, for population growth. 

\begin{figure}[hbt]
	\centering
	\includegraphics[width = .9\textwidth]{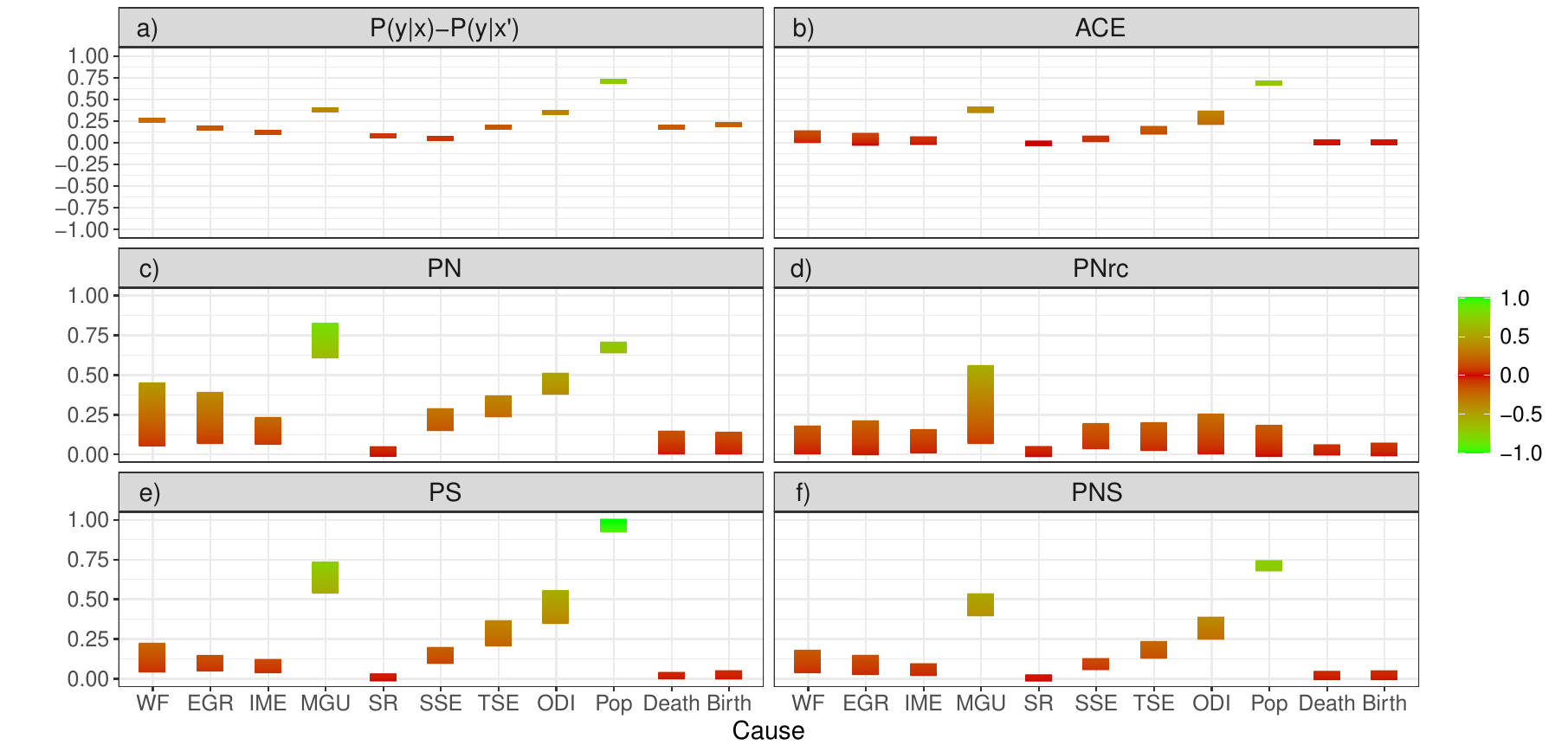}
	\caption{Intervals for the queries with $Built$ as effect variable.}
	\label{fig:BUILT}
\end{figure}

The remaining \textit{effect} variables are all referred to land-uses. For variable \textit{Built} which represents the percentage of built or artificial areas in a municipality, only two variables turn out to have a significant causal effect, namely the location of the municipality, \textit{MGU} and its population density, \textit{Pop}. It is no wonder that \textit{Pop} is the most clear sufficient cause of positive value of \textit{Built}, with $\mathrm{PS}(Pop,Built)> 0.93$ (Figure~\ref{fig:BUILT}~(e)). However, the probability of necessity for this variable is lower, but still remarkable, between $0.65$ and $0.70$ (Figure~\ref{fig:BUILT}~(c)). The same probability reaches higher values for the location of the municipality, reaching the interval $[0.62,0.82]$. Nonetheless, the probability of sufficiency of \textit{MGU} is in $[0.55,0.73]$, which is notably lower than that for the other variable. 
Moreover, both causes have a good chance of being necessary and sufficient for \textit{Built} to be abundant or fairly abundant (Figure~\ref{fig:BUILT}~(f)).
Hence, the results of the queries seem to indicate that the location (flat vs. mountainous areas) is fundamental when determining the percentage of built area, but also in combination with the population density \citep{ehrlich2021,thornton2022}.

\begin{figure}[hbt]
	\centering
	\includegraphics[width = .9\textwidth]{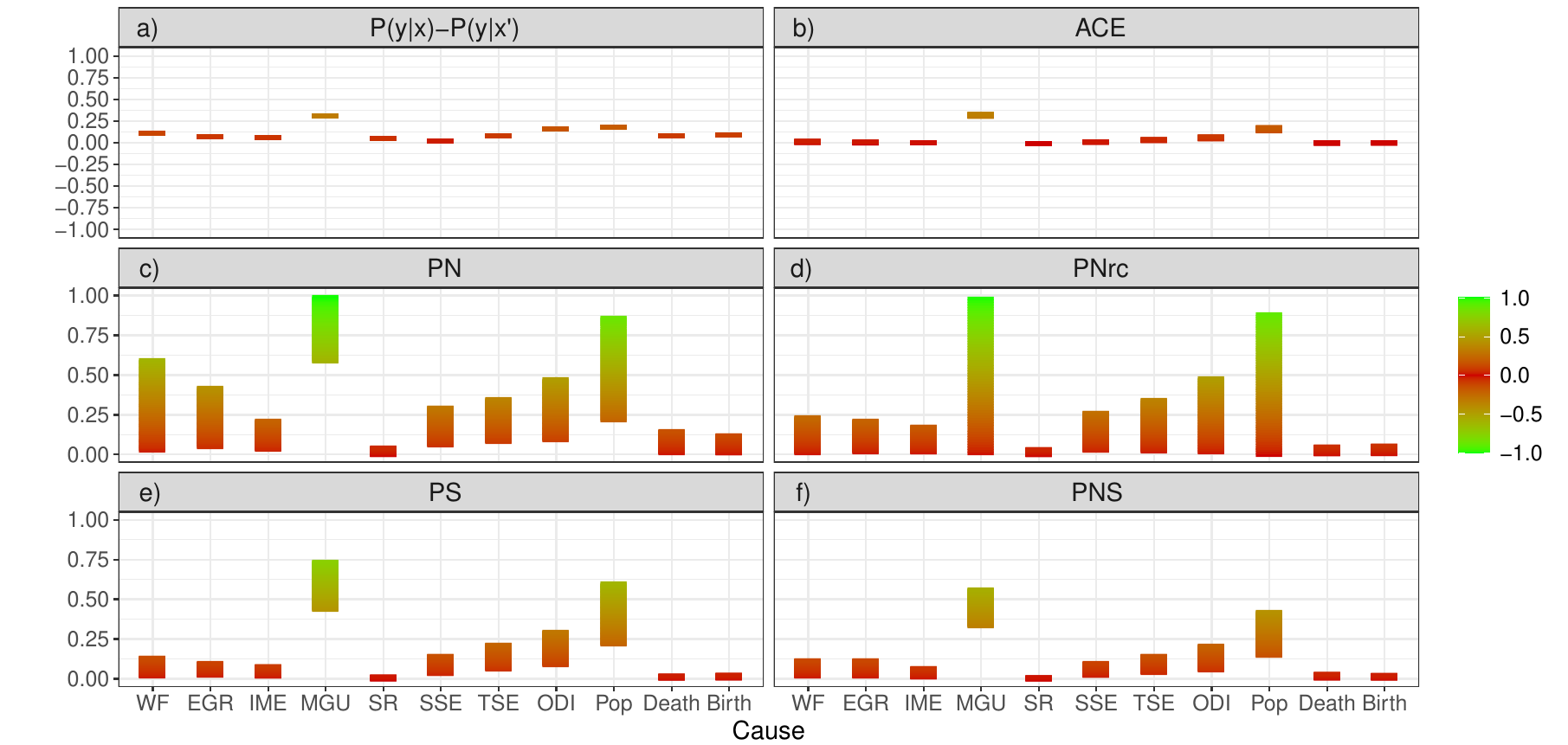}
	\caption{Intervals for the queries with $HCrops$ as effect variable.}
	\label{fig:HERBCROPS}
\end{figure}

Figure~\ref{fig:HERBCROPS} shows the results of the queries related to variable \textit{HCrops}, which is the percentage of herbaceous crops in the municipality. In this case, only variable \textit{MGU} reaches a value clearly above $0.5$, and only for one query, the probability of necessity, which is estimated to be in the interval $[0.59,0.99]$ (Figure~\ref{fig:HERBCROPS}~(c)). It means that it is quite likely that a positive value of \textit{MGU} (littoral, Baetic depression) is necessary for a positive value of $HCrops$ (dominant, fair) to be observed, but it is not so clearly rendered as sufficient too, since $\mathrm{PS}(MGU,HCrops) \in [0.44,0.74]$ (Figure~\ref{fig:HERBCROPS}~(e)), however there is still some chance that it is sufficient.
On the other hand, \textit{WF} and \textit{Pop} show a high uncertainty with respect to the probability of necessity, since their intervals are considerably wide (Figure~\ref{fig:HERBCROPS}~(c)). The remaining variables are clearly not necessary, nor sufficient, for \textit{HCrops}. 
The findings align coherently with the insights derived from Figure~\ref{fig:maps}~(c). Herbaceous crops emerge as the predominant land-use across a significant portion of the Baetic Depression, whereas they do not hold the same prevalence in other geomorphological units \citep{molero2017}.

\begin{figure}[hbt]
	\centering
	\includegraphics[width = 0.9\textwidth]{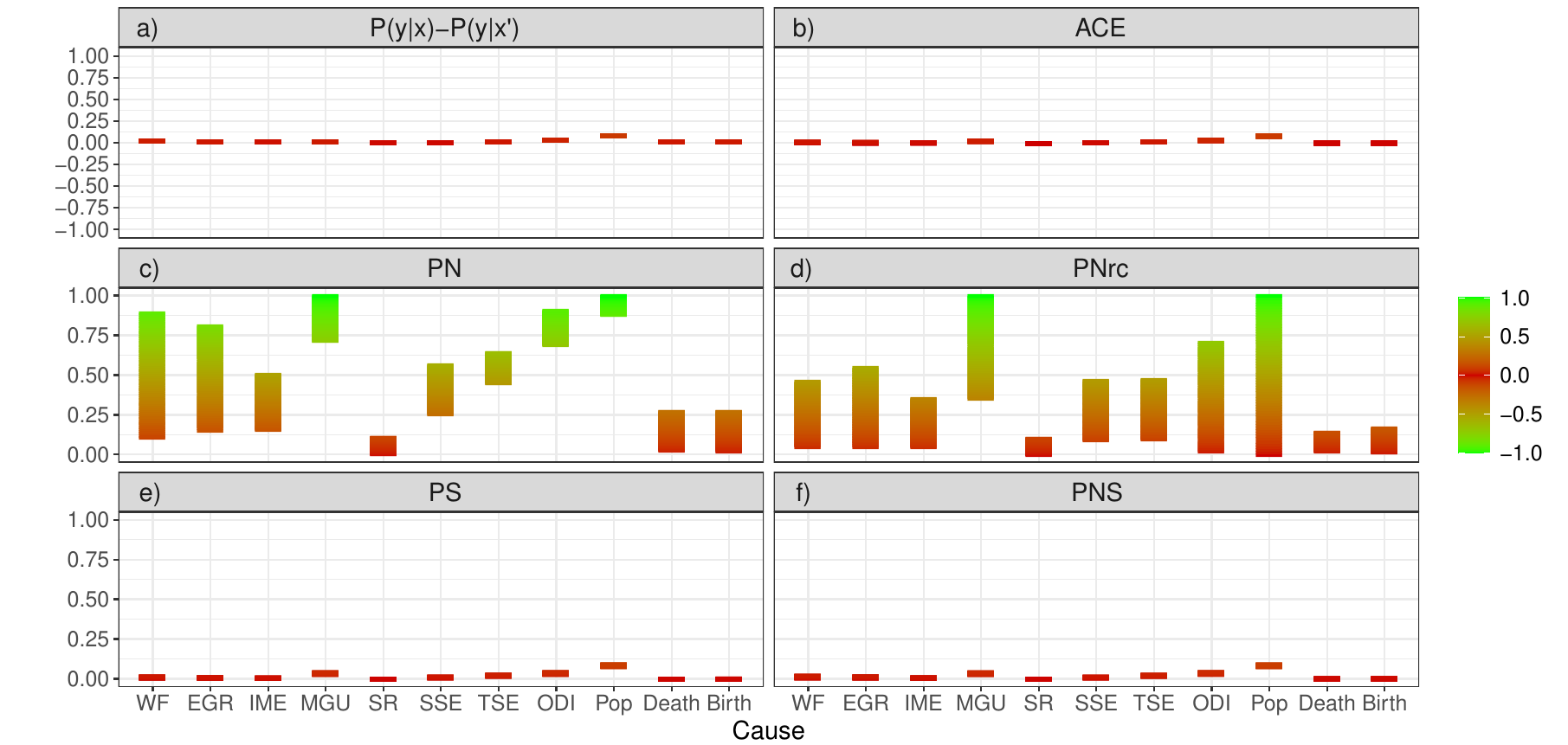}
	\caption{Intervals for the queries with $GH$ as effect variable.}
	\label{fig:INTENSIVE}
\end{figure}

In the case of the percentage of greenhouses (variable \textit{GH}), the results are displayed in Figure~\ref{fig:INTENSIVE}. It is apparent from the plots that only three variables, \textit{MGU}, \textit{Pop} and \textit{ODI}, have a causal impact on the target variable \textit{GH} (Figure~\ref{fig:INTENSIVE}~(c)). The most remarkable effect is observed in terms of the probability of necessity, with values in the intervals $[0.72,1]$, $[0.88,1]$ and $[0.69,0.91]$ for \textit{MGU}, \textit{Pop} and \textit{ODI}, respectively, which indicates that it is considered almost certain that all three variables must have a positive value for \textit{GH} to have a positive value as well \citep{aznar2011,mendoza2021}. In addition, variable \textit{TSE}, with the upper bound for PN above $0.5$ and the lower one close to $0.5$, has a good chance to be necessary \citep{galdeano2013}. It is also quite significant that none of the variables is sufficient condition by its own (Figure~\ref{fig:INTENSIVE}~(e)). With respect to the non-counterfactual queries, none of the insights previously mentioned are reflected in such queries (Figure~\ref{fig:INTENSIVE}~(a-b)), which showcases the advantage of counterfactual reasoning. The results coherently reflect the predominant location of greenhouses in the Littoral (Figure~\ref{fig:maps}~(c)). However, it is important to note that the positive state of \textit{MGU} encompasses both the Littoral and the Baetic Depression, with the latter not exhibiting a notably high density of greenhouses. Therefore, \textit{MGU} is necessary but not sufficient condition for \textit{GH} \citep{wolosin2008}.

\begin{figure}[hbtp]
	\centering
	\includegraphics[width = .9\textwidth]{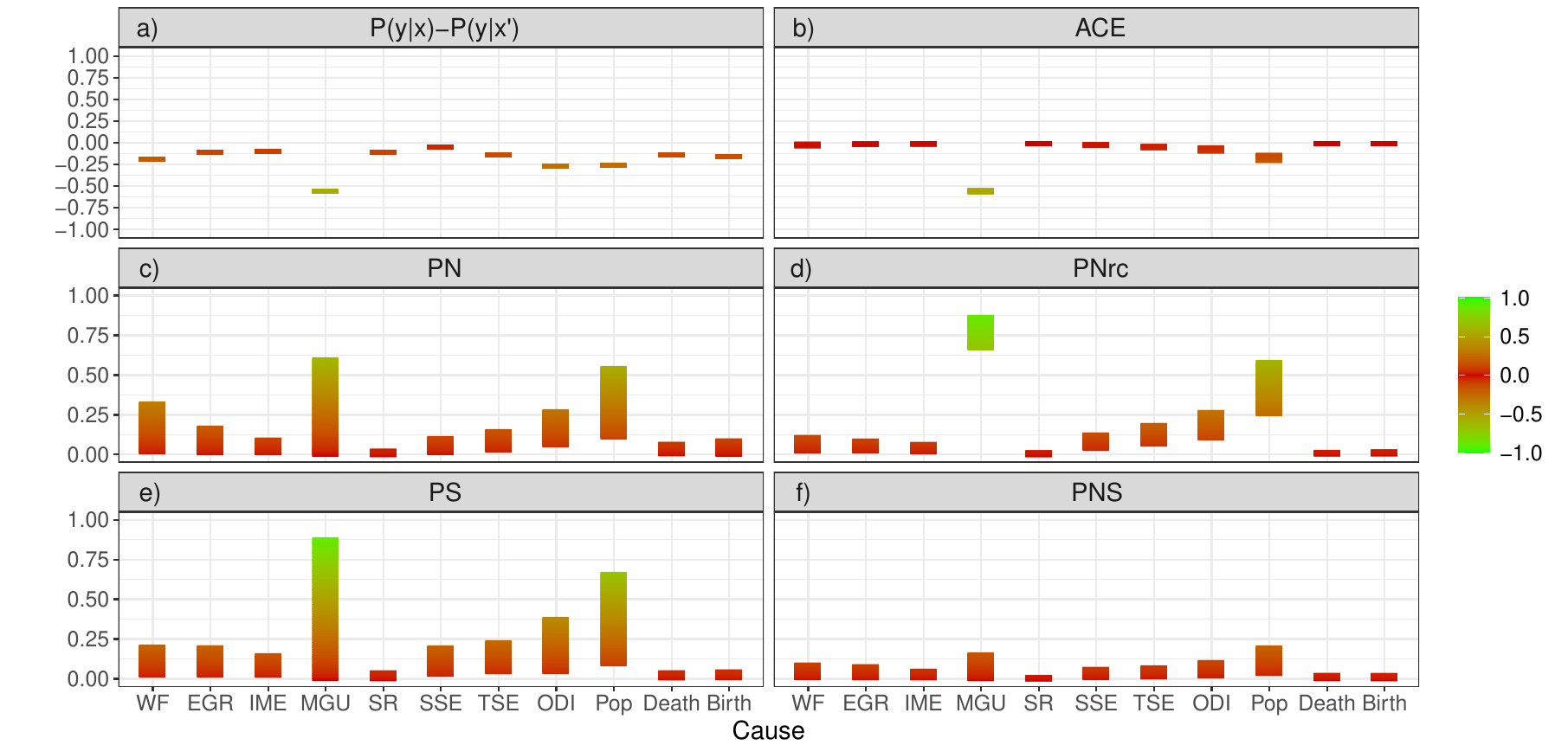}
	\caption{Intervals for the queries with $Natural$ as effect variable.}
	\label{fig:Natural}
\end{figure}

The results for variable \textit{Natural}, representing the percentage of natural areas, can be seen in Figure~\ref{fig:Natural}. In terms of the difference in conditional probability and average causal effect, only the location (\textit{MGU}) shows a remarkable variation (Figure~\ref{fig:Natural}~(a-b)). Regarding the probability of necessity, all variables except \textit{MGU} and \textit{Pop} are very unlikely to be necessary, due to the low value of the upper bounds of their intervals. Even the two mentioned variables are not clearly necessary, since most of their intervals are below $0.5$, and their widths indicate uncertainty about the probability value (Figure~\ref{fig:Natural}~(c)). 
It is also clear that all variables except \textit{MGU} and \textit{Pop} are very unlikely to be sufficient to increase the natural spaces. The intervals for \textit{MGU} and \textit{Pop} are compatible with them being sufficient conditions, but their amplitude indicate that there is considerable uncertainty (Figure~\ref{fig:Natural}~(e)). 
The most remarkable insight about the natural areas can be obtained from the high value in the query PNrc with \textit{MGU} as cause variable (Figure~\ref{fig:Natural}~(d)). This shows that not being in the littoral nor in the Baetic depression is necessary to have mainly a natural land-use, since $\mathrm{PNrc}(MGU, Natural)\in [0.67,0.87]$. In other words, mountainous areas (Sierra Morena and Baetic Systems) are necessary for the natural areas to be dominant over the other land-uses.
The results align with the fact that natural lands are predominantly located in the Baetic Systems and Sierra Morena mountain ranges, as shown in Figure~\ref{fig:maps}~(c) \citep{gratzer2017,snethlage2022}.

\begin{figure}[h!]
	\centering
	\includegraphics[width = .9\textwidth]{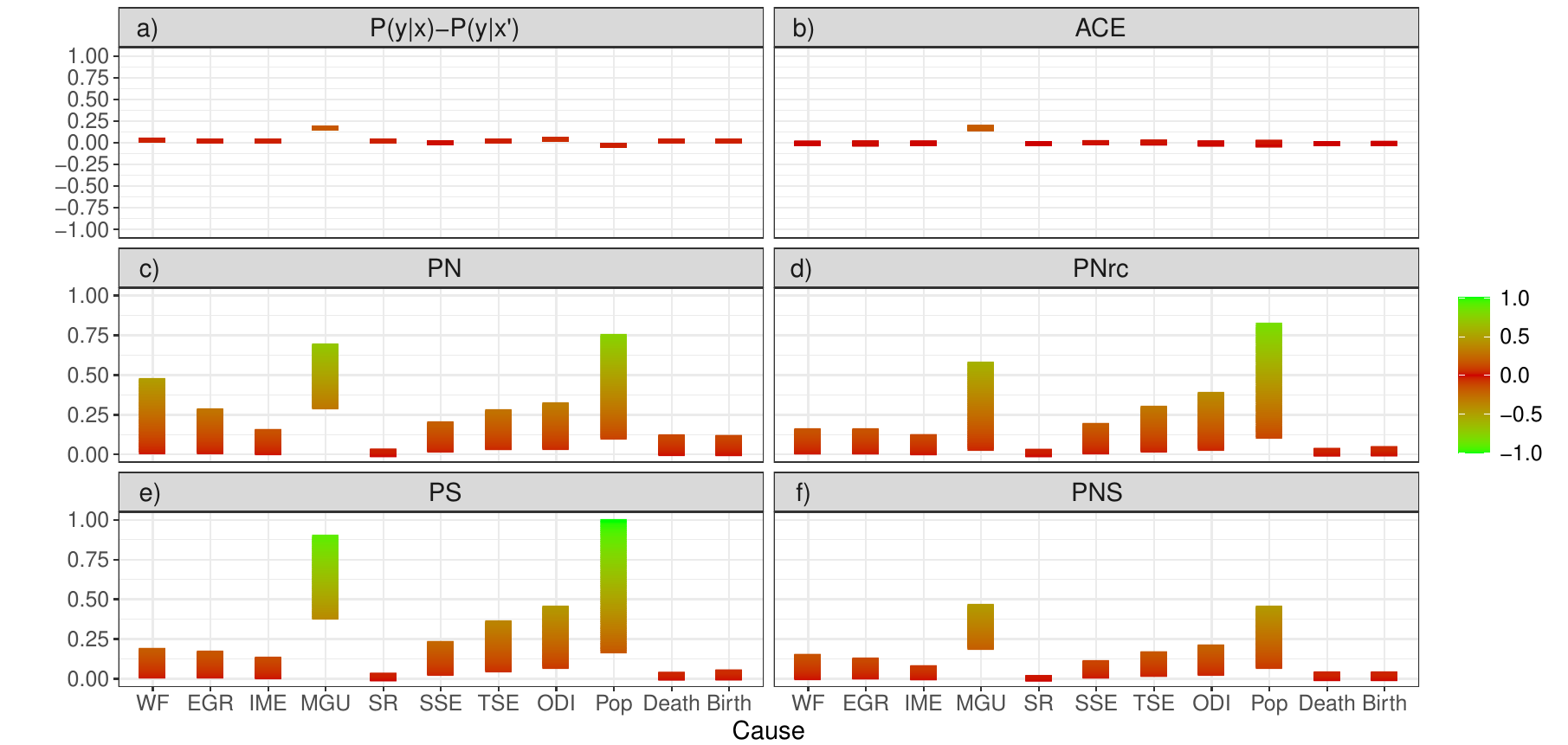}
	\caption{Intervals for the queries with $WCrops$ as effect variable.}
	\label{fig:WoodCrops}
\end{figure}

Figure~\ref{fig:WoodCrops} shows the results for variable \textit{WCrops}, which measures the percentage of woody crops in each municipality. It can be seen that none of the variables shows a significant variation in conditional probability or remarkable values of ACE (Figure~\ref{fig:WoodCrops}~(a-b)). Considering the queries related to necessity and sufficiency, \textit{MGU} and \textit{Pop} might have a significant influence on \textit{WCrops}, but the width of the corresponding intervals poses a high level of uncertainty on such statement (Figure~\ref{fig:WoodCrops}~(c-f)). The rest of possible causes are clearly not relevant.

\begin{figure}[hbt]
	\centering
	\includegraphics[width = .9\textwidth]{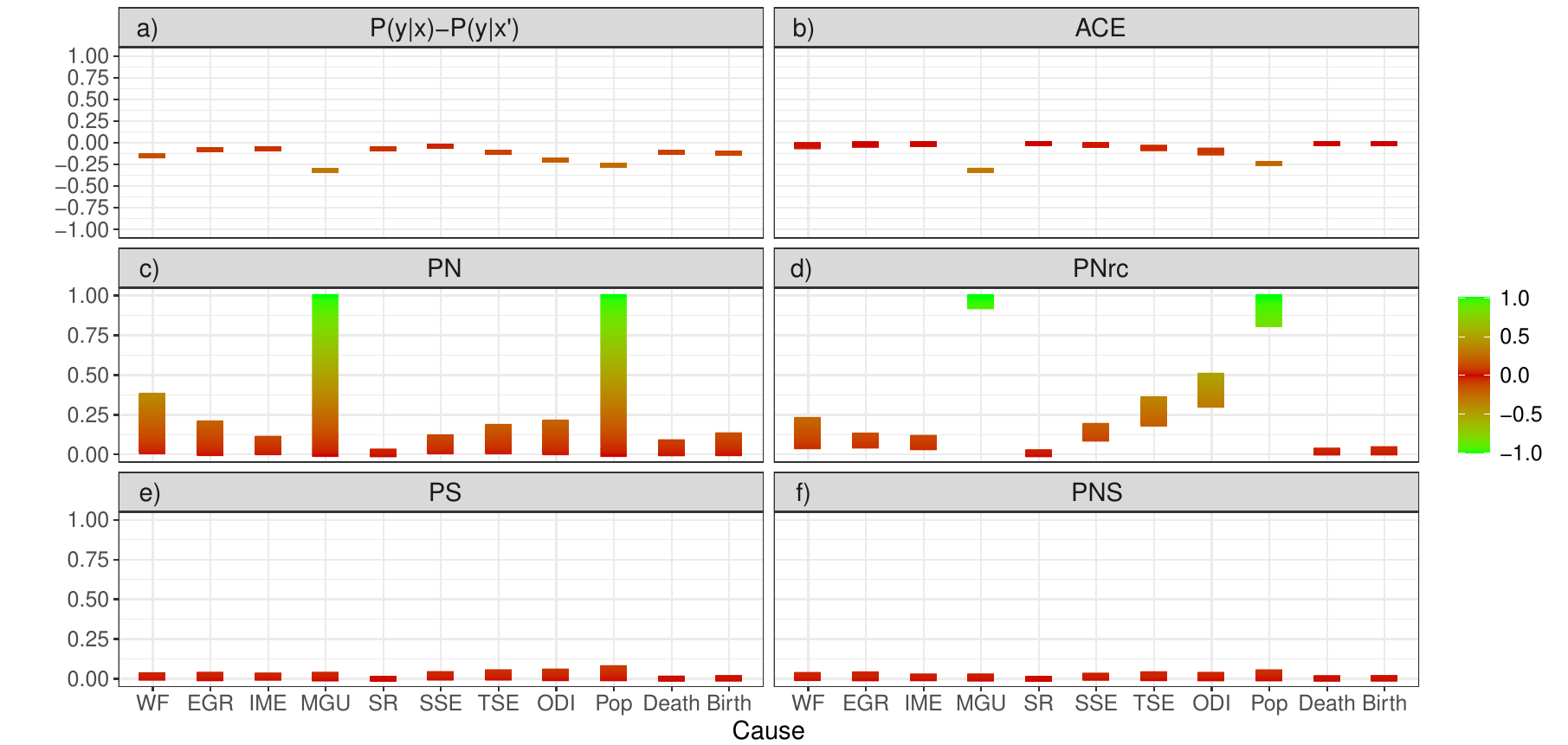}
	\caption{Intervals for the queries with $Mixed$ as effect variable.}
	\label{fig:HETEROGENEO}
\end{figure}

Finally, Figure~\ref{fig:HETEROGENEO} displays the results for variable \textit{Mixed} (percentage of heterogeneous lands). The only causal effect in terms of difference in conditional probability and ACE is provided by variables \textit{Pop} and \textit{MGU} (Figure~\ref{fig:HETEROGENEO}~(a-b)). It is highly unlikely that all variables except \textit{MGU} and \textit{Pop} are necessary (see PN, PNrc and PNS in Figure~\ref{fig:HETEROGENEO}~(c), (d) and (f), respectively), but even for these two variables there is paramount uncertainty in terms of PN. However, PNrc clearly points towards the facts that not being in the littoral nor in the Baetic depression and having a low population density are both necessary conditions. On the other hand, the values of PS and PNS clearly show that none of the variables are sufficient for \textit{Mixed} to have a positive value (Figure~\ref{fig:HETEROGENEO}~(e-f)). 
The results are consistent with the fact that heterogeneous lands are the main land-use in some areas of Sierra Morena \citep{munoz2011,plieninger2021}, coinciding with a concurrent trend of population decline in those regions, as shown in Figure~\ref{fig:maps}~(c,d) \citep{plieninger2001}.

\section{Conclusions}\label{sec:conclusions}
This paper proposes the application of counterfactual reasoning with BN for analyzing socioecological systems. This approach tackles a limitation of traditional probabilistic analysis, which cannot determine the nature of the relation between variables (whether it is one of necessity or sufficiency). To address this, we suggest employing a recently developed technique based on the well-known EM algorithm for parametric learning in BNs with latent variables. An advantage of this method is its capability to calculate bounds for certain queries, known as unidentifiable. Notably, this is achieved through the analysis of observational data alone, unlike other methods that also require interventional data. 
Note that the primary requirement for adopting this framework is the definition of the causal graph, which can be specified by experts, as is the case in the current study. However, if the graph involves many connections per node, the number of parents may need to be limited to reduce computational complexity.
The model parameters are then learned from observational data only.

To demonstrate the utility of counterfactual reasoning, we have presented a case study using an observational dataset containing information on socioeconomic factors and land-uses in southern Spain. 
The primary conclusion drawn from this study is that both the location and population density are essential for most land-uses. Specifically, a mountainous location with low population density is deemed necessary for natural or mixed land-uses. Conversely, a non-mountainous location with high population density is required for the presence of built areas, herbaceous crops, or greenhouses. 
Concerning population dynamics, our study indicates that immigration not only is necessary but also sufficient for population growth. All these findings underscores the power of counterfactual reasoning in uncovering relationships within socioecological systems, in contrast to the plain use of BNs for which only observational queries can be solved.

While this study provides valuable insights, further research could be carried out to consider counterfactual queries with multiple cause variables, for instance to determine whether simultaneous causes are jointly sufficient. 
Additionally, this methodology could be of considerable interest for other topics within the environmental and ecological areas, such as the study of species distribution or risk assessments.


\section*{Code and data availability} The code and data-set used in the current study is available at the repository at \url{https://github.com/PGM-Lab/2023-counterfactual-land}. 

\section*{CRediT authorship contribution statement} Conceptualization, R.C., A.D.M. and P.A.A.; methodology R.C; software R.C. and M.M.; data curation A.D.M. and M.M.;formal analysis R.C.,  A.D.M. and A.S.; writing–original draft preparation, R.C., A.D.M., M.M., P.A.A. and A.S.; funding acquisition, A.S.; supervision, P.A.A. and A.S. All authors have read and agreed to the published version of the manuscript.

\section*{Declaration of generative AI and AI-assisted technologies in the
writing process}
The authors did not use any form of generative AI during the preparation
of this work. The writing and editing process was done by human
hand.

\section*{Funding}

Grant PID2022-139293NB-C31 funded by MCIN/AEI/10.13039/501100011033 and by ERDF A way of making Europe. 
R.C. acknowledges the support by Spanish Ministry of Science, Innovation and Universities through the ``Mar\'{i}a Zambrano'' grant (RR\_C\_2021\_01) funded with NextGenerationEU funds. Rafael Cab\~{n}as was also supported by ``Plan Propio de Investigaci\'{o}n y Transferencia 2024-2025''  from University of Almer\'{i}a under the project P\_LANZ\_2024/003. 

\section*{Declaration of competing interest} The authors have no relevant financial or non-financial interests to disclose.

\appendix
\section{Structural causal models and Bayesian networks}
\label{ap:scm}
Structural causal models (SCMs)~\citep{pearl2009} are a specific type of probabilistic graphical model used for causal and counterfactual reasoning. SCMs can be formally defined as follows \citep{bareinboim2022pearl}.\newline

\begin{definition}[Structural causal model (SCM)] A structural causal model $\mcM$ is a 4-tuple $\langle \bmU, \bmX,\mcF_{\bmX}, \mcP_{\bmU} \rangle$, where
	
	\begin{itemize}
		\item $\bmU$ is a set of exogenous variables that are determined by factors outside the model;
		\item $\bmX$ is a set of variables $\{X_1, X_2,\ldots,X_n\}$, called endogenous, that are determined by other (exogenous and endogenous) variables in the model,  i.e. by variables in $\bmU \cup \bmX$. 
		\item $\mcF_{\bmX}$ is a set of functions $\{f_{X_1}, f_{X_2},..., f_{X_n}\}$ called \textit{structural equations} (SE), such that each of them is a function $f_{X_i}:\Omega_{\bmU_i}\cup \Omega_{\mathrm{Pa}_{X_i}} \to \Omega_{X_i}$, where $\mathrm{Pa}_{X_i} \subseteq \bmX$ are the endogenous variables directly determining $X_i$ and $\bmU_i\subseteq \bmU$ are the exogenous variables directly determining $X_i$. 
		\item $\mcP_{\bmU}$ is a set containing a probability distribution $P(U)$ for each $U\in\bmU$. 
	\end{itemize}
	
\end{definition}

Note that the structural equations $\mcF_{\bmX}$ actually define a directed acyclic graph (DAG) $\mcG$ called the \emph{causal graph} of the model, whose nodes correspond to the variables in $\bmU \cup \bmX$ and containing a link from each variable in ${\bmU_i}\cup {\mathrm{Pa}_{X_i}}$ to $X_i$, $i=1,\ldots,n$.

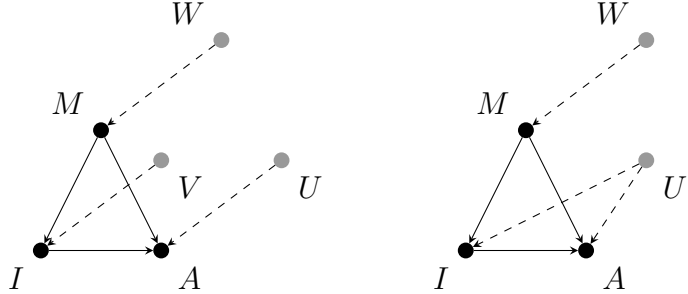
\begin{figure}[htp!]
	\centering
	\begin{tabular}{cp{10pt}c}
		\begin{tikzpicture}[scale=0.8]
			\node[dot,label=above left:{$M$}] (M)  at (1,1) {};
			\node[dot,label=below left:{$I$}] (I)  at (0,-1) {};
			\node[dot,label=below right:{$A$}] (A)  at (2.0,-1) {};
			\node[dot2,label=above left:{$W$}] (W)  at (3,2.5) {};
			\node[dot2,label=below right:{$V$}] (V)  at (2,0.5) {};
			\node[dot2,label=below right:{$U$}] (U)  at (4,0.5) {};
			\draw[a2] (M) -- (I);
			\draw[a2] (M) -- (A);
			\draw[a2] (I) -- (A);
			\draw[a] (W) -- (M);
			\draw[a] (V) -- (I);
			\draw[a] (U) -- (A);
		\end{tikzpicture}
		& &
		\begin{tikzpicture}[scale=0.8]
			\node[dot,label=above left:{$M$}] (M)  at (1,1) {};
			\node[dot,label=below left:{$I$}] (I)  at (0,-1) {};
			\node[dot,label=below right:{$A$}] (A)  at (2.0,-1) {};
			\node[dot2,label=above left:{$W$}] (W)  at (3,2.5) {};
			\node[dot2,label=below right:{$U$}] (U)  at (3,0.5) {};
			\draw[a2] (M) -- (I);
			\draw[a2] (M) -- (A);
			\draw[a2] (I) -- (A);
			\draw[a] (W) -- (M);
			\draw[a] (U) -- (I);
			\draw[a] (U) -- (A);
		\end{tikzpicture}
	\end{tabular}
	\caption{Examples of two possible causal graphs for the problem shown in Figure~\ref{fig:into_bn}. $U, V$ and $W$ are exogenous variables.}\label{fig:example_cgraph}
\end{figure}

As an illustrative example, we will begin by examining Figure~\ref{fig:example_cgraph}, which depicts two potential causal graphs for SCMs extending the BN presented in Figure~\ref{fig:into_bn}.  The endogenous variables, represented as black nodes, correspond to the variables originally found in the initial BN, specifically $\bmX = \{M, I, A\}$. These variables retain their original domains. In addition to the endogenous variables, the causal graphs also incorporate exogenous variables, which are depicted as gray nodes. In the left graph, the set of exogenous variables is $\bmU = \{U, V, W\}$, while in the right graph, it is $\bmU = \{U, W\}$.

\newcommand\independent{\protect\mathpalette{\protect\independenT}{\perp}}
\def\independenT#1#2{\mathrel{\rlap{$#1#2$}\mkern2mu{#1#2}}}

The distinction between the two causal graphs is as follows: the graph on the left assumes the absence of any exogenous (and hidden) confounder between any pair of variables. In contrast, in the graph on the right, variables $I$ and $A$ are both influenced by the exogenous variable $U$. According to the classification given by \cite{avin2005identifiability}, a SCM with a graph like the one on the left is termed \textit{Markovian}, meaning that each exogenous variable has only one endogenous child. Conversely, a SCM with a graph like the one on the right is termed \textit{semi-Markovian}, indicating that any of the exogenous variables can have more than one endogenous child\footnote{Some authors consider a less general definition and limit exogenous variables in semi-Markovian models to have no more than 2 children \citep{huang2006identifiability}.}. It is important to note that in both cases, all endogenous variables must have exactly one exogenous parent. In other words, given the endogenous variables, each exogenous variable is independent of the others, which is denoted as $U_i \independent U_j | \bmX$ for all $U_i,U_j \in \bmU\times\bmU$ with $i\neq j$. If this condition is not satisfied the model will be classified as \textit{non-Markovian}.

A parameterization for the previously mentioned SCM is illustrated in Figure~\ref{fig:scm_params}. The set of marginal distributions associated with the exogenous variables is $\mcP_{\bmU} = \{P(U), P(V), P(W)\}$. Conversely, the set of SEs is represented by $\mcF_{\bmX} = \{f_A(U, I, M), f_I(V, M), f_M(W)\}$. 

\begin{figure}[htp!]
	\centering
	\begin{tikzpicture}[scale=0.8]
		\node[scale=0.85] at (0, 0) {$
			P(W)=
			\begin{blockarray}{cc}
				\color{gray}{w_1} &  \color{gray}{w_2}  \\
				\begin{block}{[cc]}
					0.7253 & 0.2747\\
				\end{block}
			\end{blockarray}
			$};
		
		\node[scale=0.85] at (8.3, 0) {$
			P(V)=
			\begin{blockarray}{cccc}
				\color{gray}{v_1} &  \color{gray}{v_2} & \color{gray}{v_3} &  \color{gray}{v_4} \\
				\begin{block}{[cccc]}
					0.56312 & 0 & 0.19565 & 0.24123\\
				\end{block}
			\end{blockarray}
			$};

		\node[scale=0.85] at (4.9, -1.4) {$
			P(U)=
			\begin{blockarray}{cccccccccc}
				\color{gray}{u_1} &  \color{gray}{u_2} & \color{gray}{u_3} &  \color{gray}{u_4} & \color{gray}{u_5} &  \color{gray}{u_6} & \color{gray}{u_7} &  \color{gray}{u_8} & \color{gray}{u_9} &  \color{gray}{u_{10}} \\
				\begin{block}{[cccccccccc]}
					0 & 0.24979 & 0 & 0 & 0.03045 & 0.30418 & 0 & 0.14545 & 0 & 0.27013\\
				\end{block}
			\end{blockarray}
			$};

		\node[scale=0.85] at (1, -3.0) {$
			f_M(W)=
			\begin{blockarray}{cc}
				\color{gray}{w_1} &  \color{gray}{w_2} \\
				\begin{block}{[cc]}
					yes & no\\
				\end{block}
			\end{blockarray}
			$};
		
		\node[scale=0.85] at (8, -3.0) {$
			f_I(M,V)=
			\begin{blockarray}{ccccl}
				\color{gray}{v_1} &  \color{gray}{v_2} & \color{gray}{v_3} &  \color{gray}{v_4} &\\
				\begin{block}{[cccc]l}
					yes & yes& no & no &\tiny{\color{gray}{\tiny M = yes} }\\
					yes & no & yes & no & {\color{gray}{M = no} }\\
				\end{block}
			\end{blockarray}
			$};

		\node[scale=0.82] at (5, -5.5) {$
			f_A(M,I,U)=
			\begin{blockarray}{ccccccccccl}
				\color{gray}{u_1} &  \color{gray}{u_2} & \color{gray}{u_3} &  \color{gray}{u_4} & \color{gray}{u_5} &  \color{gray}{u_6} & \color{gray}{u_7} &  \color{gray}{u_8} & \color{gray}{u_9} &  \color{gray}{u_{10}} &\\
				\begin{block}{[cccccccccc]l}
					yes & yes & yes & yes & yes & no & no & no & no & no &\tiny{\color{gray}{\tiny M = yes, I=yes} }\\
					yes & no& yes & yes & no & yes & no & no & yes & no & \tiny{\color{gray}{\tiny M = yes, I=no} }\\
					yes & yes& no & no & no & yes & yes & yes & no & no & \tiny{\color{gray}{\tiny M = no, I=yes} }\\
					yes & yes & yes & no & yes & yes & yes & no & yes & yes & \tiny{\color{gray}{\tiny M = no, I=no} }\\
				\end{block}
			\end{blockarray}
			$};

	\end{tikzpicture}
	\caption{SEs and marginal distributions for the Markovian SCM in Figure~\ref{fig:example_cgraph} (left).}\label{fig:scm_params}
\end{figure}
In the formalism of SCMs, SEs are typically assumed to be provided, often derived from expert knowledge. Alternatively, SEs can be automatically inferred from the causal graph, without any loss of generality, via \emph{canonical specification} \citep{zhang2022partial}. The states of an exogenous variable will then represent all possible function mappings between  its children domains from their respective endogenous parents domains. 
In this sense, an exogenous variable in a Markovian model, with $X$ as its child, would require the number of states given by the following expression. 

\begin{equation}
|\Omega_U| =
\begin{cases}
|\Omega_X|, & \text{if } Pa_X = \emptyset \\
|\Omega_X|^{|\Omega_{Pa_X}|}, & \text{otherwise}
\end{cases}
\label{eq:omega_U}
\end{equation}

In Figure~\ref{fig:scm_params}, the SEs associated with variables $M$ and $I$ are already canonical, whereas the one for $A$ is not, since the number of possible values of $U$ (10 in this case) does not match $|\Omega_{A}|^{|\Omega_{Pa_A}|}$ (16 in this case). The SE under the canonical specification for $A$ and the corresponding distribution $P(U)$ are shown  in Figure~\ref{fig:scm_canonical}. \newline 

\begin{figure}[htp!]
	\centering
	\begin{tikzpicture}[scale=0.78]
		\node[scale=0.78, anchor=west] at (-0.6, 0) {$
			P(U)=
			\begin{blockarray}{cccccccccccccccc}
				\color{gray}{u_1} &  \color{gray}{u_2} & \color{gray}{u_3} &  \color{gray}{u_4} & \color{gray}{u_5} &  \color{gray}{u_6} & \color{gray}{u_7} &  \color{gray}{u_8} & \color{gray}{u_9} &  \color{gray}{u_{10}} &  \color{gray}{u_{11}} & \color{gray}{u_{12}} &  \color{gray}{u_{13}} & \color{gray}{u_{14}} &  \color{gray}{u_{15}} & \color{gray}{u_{16}} \\
				\begin{block}{[cccccccccccccccc]}
					0 & 0 & 0 & 0 & 0.16 & 0.12 & 0 & 0 & 0 & 0 & 0.02 & 0.68 & 0 & 0 & 0 & 0.02 \\
				\end{block}
			\end{blockarray}
			$};

		\node[scale=0.78, anchor=west] at (-0.6, -1) {
			$
			f_A(M,I,U)=
			$
		};
		\node[scale=0.68, anchor=west] at (-0.6, -3.0) {
			\parbox{\textwidth}{
				$
				\begin{blockarray}{ccccccccccccccccl}
					\color{gray}{u_1} &  \color{gray}{u_2} & \color{gray}{u_3} &  \color{gray}{u_4} & \color{gray}{u_5} &  \color{gray}{u_6} & \color{gray}{u_7} &  \color{gray}{u_8} & \color{gray}{u_9} &  \color{gray}{u_{10}} &  \color{gray}{u_{11}} & \color{gray}{u_{12}} &  \color{gray}{u_{13}} & \color{gray}{u_{14}} &  \color{gray}{u_{15}} & \color{gray}{u_{16}} \\
					\begin{block}{[cccccccccccccccc]l}
						yes & yes & yes & yes & yes & yes & yes & yes & no & no & no & no & no & no & no & no &\tiny{\color{gray}{\tiny  M=yes, I=yes}}\\
						yes & yes& no& no&yes &yes &no &no & yes&yes &no &no & yes&yes & no&no &\tiny{\color{gray}{\tiny   M=yes, I=no}}\\
						yes & yes& yes&yes &no &no &no &no &yes & yes&yes & yes&no &no & no&no &\tiny{\color{gray}{\tiny   M=no, I=yes}}\\
						yes &no & yes& no& yes& no& yes& no& yes& no&yes & no& yes& no& yes& no&\tiny{\color{gray}{\tiny   M=no, I=no}}\\
					\end{block}
				\end{blockarray}
				$}
		};
	\end{tikzpicture}
	\caption{Canonical SE for $A$ and marginal distribution for exogenous variable $U$ in the Markovian SCM in Figure~\ref{fig:example_cgraph}.}\label{fig:scm_canonical}
\end{figure}

As already considered in some works from the literature \citep{zaffalon2020structural,zaffalon2023efficient}, a SCM can be specified as a BN as follows. First, the graphical component is the same: the causal graph in the SCM is the DAG $\mcG$ in the BN. In this way, the BN is defined over the union of the exogenous and endogenous sets of nodes, i.e. $\bmV = \bmU \cup \bmX$. Regarding the distributions in $\mcP_{\bmV}$, each exogenous variable is associated with the corresponding marginal distribution in $\mcP_{\bmU}$, whereas each SE $f_X(Pa_X)$ induces a CPT of the form $P(X|Pa_X)$, characterized by containing only ones and zeros. More precisely, for each $(x,\pi_X) \in \Omega_{X}\times\Omega_{Pa_X}$, $P(x|\pi_X)$ takes the value 1 if $f_X(\pi_X) = x$ and 0 otherwise. For instance, Figure~\ref{fig:bn_scm_params} shows the SEs from the running example represented as CPTs.

\begin{figure}[htp!]
	\centering
	\begin{tikzpicture}[scale=0.8]
		\node[scale=0.85] at (-2.2, -3.0) {$
			P(M|W)=
			\begin{blockarray}{ccl}
				\color{gray}{w_1} &  \color{gray}{w_2} & \\
				\begin{block}{[cc]l}
					1 & 0& {\color{gray}{M = yes} }\\
					0 & 1& \color{gray}{M = no} \\
				\end{block}
			\end{blockarray}
			$};
		
		\node[scale=0.85] at (6, -3.0) {$
			P(I|M,V)=
			\begin{blockarray}{ccccl}
				\color{gray}{v_1} &  \color{gray}{v_2} & \color{gray}{v_3} &  \color{gray}{v_4} &\\
				\begin{block}{[cccc]l}
					1 & 1& 0 & 0 &\tiny{\color{gray}{\tiny M = yes, I=yes} }\\
					0 & 0& 1 & 1 &\color{gray}{M = yes, I=no} \\
					1 & 0 & 1 & 0 & {\color{gray}{M = no, I=yes} }\\
					0 & 1 & 0 & 1 & \color{gray}{M = no, I=no} \\
				\end{block}
			\end{blockarray}
			$};

		\node[scale=0.80] at (3, -7) {$
			P(A|M,I,U)=
			\begin{blockarray}{ccccccccccl}
				\color{gray}{u_1} &  \color{gray}{u_2} & \color{gray}{u_3} &  \color{gray}{u_4} & \color{gray}{u_5} &  \color{gray}{u_6} & \color{gray}{u_7} &  \color{gray}{u_8} & \color{gray}{u_9} &  \color{gray}{u_{10}} &\\
				\begin{block}{[cccccccccc]l}
					1 & 1 & 1 & 1 & 1 & 0 & 0 & 0 & 0 & 0 &\tiny{\color{gray}{\tiny M = yes, I=yes, A=yes} }\\
					0 & 0 & 0 & 0 & 0 & 1 & 1& 1 & 1 & 1 & \tiny{\color{gray}{\tiny M = yes, I=yes, A=no} }\\
					1 & 0& 1 & 1 & 0 & 1 & 0 & 0 & 1 & 0 & \tiny{\color{gray}{\tiny M = yes, I=no, A=yes} }\\
					0 & 1 & 0 & 0 & 1& 0 & 1 & 1 & 0 & 1 &\tiny{\color{gray}{\tiny M = yes, I=no, A=no} }\\
					1 & 1& 0 & 0 & 0 & 1 & 1 & 1 & 0 & 0 & \tiny{\color{gray}{\tiny M = no, I=yes, A=yes} }\\
					0 & 0 & 1 & 1& 1 & 0 & 0 & 0 & 1 & 1 & \tiny{\color{gray}{\tiny M = no, I=yes, A=no} }\\
					1 & 1 & 1 & 0 & 1 & 1 & 1 & 0 & 1 & 1 & \tiny{\color{gray}{\tiny M = no, I=no, A=yes} }\\
					0 & 0 & 0 & 1 & 0 & 0 & 0 & 1 & 0 & 0 &\tiny{\color{gray}{\tiny M = no, I=no, A=no} }\\
				\end{block}
			\end{blockarray}
			$};

	\end{tikzpicture}
	\caption{SEs from Figure~\ref{fig:scm_params} represented as CPTs. $P(M|W), P(I|M,V)$ and $P(A|M,I,U)$ represent the same information as SEs $f_M, f_I$ and $f_A$, respectively.}\label{fig:bn_scm_params}
\end{figure}

\section{Computing counterfactual queries in SCMs}
\label{ap:queries}

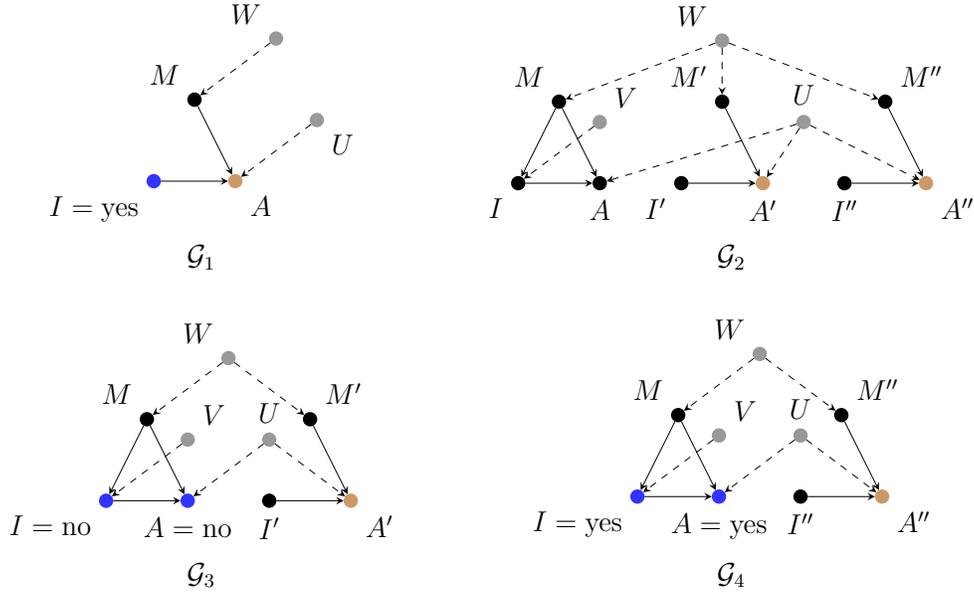
\begin{figure}[h!bt]
	\centering

        \resizebox{\textwidth}{!}{
	\begin{tabular}{cp{8pt}c}
		\begin{tikzpicture}[scale=0.6]
			\node[dot,label=above left:{$M$}] (M)  at (1,1) {};
			\node[dot3,label=below left:{$I=\text{yes}$}] (I)  at (0,-1) {};
			\node[dot4,label=below right:{$A$}] (A)  at (2.0,-1) {};
			\node[dot2,label=above left:{$W$}] (W)  at (3,2.5) {};
			\node[dot2,label=below right:{$U$}] (U)  at (4,0.5) {};
			\draw[a2] (M) -- (A);
			\draw[a2] (I) -- (A);
			\draw[a] (W) -- (M);
			\draw[a] (U) -- (A);
		\end{tikzpicture}
		& &

		\begin{tikzpicture}[scale=0.6]
  			\node[dot,label=above left:{$M$}] (M)  at (-2,1) {};
			\node[dot,label=below left:{$I$}] (I)  at (-3,-1) {};
			\node[dot,label=below:{$A$}] (A)  at (-1,-1) {};
   			\node[dot2,label=above right:{$V$}] (V)  at (-1,0.5) {};
			\node[dot,label=above left:{$M'$}] (M1)  at (2,1) {};
			\node[dot,label=above right:{$M''$}] (M2)  at (6,1) {};
			\node[dot,label=below left:{$I'$}] (I1)  at (1,-1) {};
			\node[dot,label=below :{$I''$}] (I2)  at (5,-1) {};
			\node[dot4,label=below :{$A'$}] (A1)  at (3.0,-1) {};
			\node[dot4,label=below right:{$A''$}] (A2)  at (7.0,-1) {};
			\node[dot2,label=above left:{$W$}] (W)  at (2,2.5) {};
			\node[dot2,label=above :{$U$}] (U)  at (4,0.5) {};
   			\draw[a2] (M) -- (A);
         			\draw[a2] (M) -- (I);
			\draw[a2] (I) -- (A);
			\draw[a] (W) -- (M);
			\draw[a] (U) -- (A);
   			\draw[a] (V) -- (I);
			\draw[a2] (M1) -- (A1);
			\draw[a2] (M2) -- (A2);
			\draw[a2] (I1) -- (A1);
			\draw[a2] (I2) -- (A2);
			\draw[a] (W) -- (M1);
			\draw[a] (W) -- (M2);
			\draw[a] (U) -- (A1);
			\draw[a] (U) -- (A2);
		\end{tikzpicture} \\
    ${\cal G}_1$ & & ${\cal G}_2$ \\
    & & \\
        \begin{tikzpicture}[scale=0.6]
			\node[dot,label=above left:{$M$}] (M)  at (1,1) {};
			\node[dot,label=above right:{$M'$}] (M2)  at (5,1) {};
			\node[dot3,label=below left:{$I=\text{no}$}] (I)  at (0,-1) {};
			\node[dot,label=below :{$I'$}] (I2)  at (4,-1) {};
			\node[dot3,label=below :{$A=\text{no}$}] (A)  at (2.0,-1) {};
			\node[dot4,label=below right:{$A'$}] (A2)  at (6.0,-1) {};
			\node[dot2,label=above left:{$W$}] (W)  at (3,2.5) {};
			\node[dot2,label=above :{$U$}] (U)  at (4,0.5) {};
   			\node[dot2,label=above right:{$V$}] (V)  at (2,0.5) {};
			\draw[a2] (M) -- (A);
   			\draw[a2] (M) -- (I);
			\draw[a2] (M2) -- (A2);
			\draw[a2] (I) -- (A);
			\draw[a2] (I2) -- (A2);
			\draw[a] (W) -- (M);
			\draw[a] (W) -- (M2);
			\draw[a] (U) -- (A);
			\draw[a] (U) -- (A2);
   			\draw[a] (V) -- (I);
		\end{tikzpicture}
		& &
		\begin{tikzpicture}[scale=0.6]
			\node[dot,label=above left:{$M$}] (M)  at (1,1) {};
			\node[dot,label=above right:{$M''$}] (M2)  at (5,1) {};
			\node[dot3,label=below left:{$I=\text{yes}$}] (I)  at (0,-1) {};
			\node[dot,label=below :{$I''$}] (I2)  at (4,-1) {};
			\node[dot3,label=below :{$A=\text{yes}$}] (A)  at (2.0,-1) {};
			\node[dot4,label=below right:{$A''$}] (A2)  at (6.0,-1) {};
			\node[dot2,label=above left:{$W$}] (W)  at (3,2.5) {};
			\node[dot2,label=above:{$U$}] (U)  at (4,0.5) {};
   			\node[dot2,label=above right:{$V$}] (V)  at (2,0.5) {};
			\draw[a2] (M) -- (A);
   			\draw[a2] (M) -- (I);
			\draw[a2] (M2) -- (A2);
			\draw[a2] (I) -- (A);
			\draw[a2] (I2) -- (A2);
			\draw[a] (W) -- (M);
			\draw[a] (W) -- (M2);
			\draw[a] (U) -- (A);
			\draw[a] (U) -- (A2);
   			\draw[a] (V) -- (I);
		\end{tikzpicture} \\
    ${\cal G}_3$ & & ${\cal G}_4$ 
	\end{tabular}
 }
	\caption{Graphs of the post-interventional and counterfactual models for calculating various interventional and counterfactual queries in the SCM from Figure~\ref{fig:example_cgraph} (left). Observed variables and target variables are shown in blue and yellow, respectively.}\label{fig:graph_queries}
\end{figure}

While observational queries can be calculated directly in the original model, interventional and counterfactual queries require applying graphical operations in the causal graph. Figure~\ref{fig:graph_queries} depicts the modified graphs for various queries in the motivational example. Interventional queries are  calculated in the so-called \textit{post-intervention model}, which is the result of applying a graphical operation involving the removal of incoming arcs into the intervened variable and the replacement of its SE with a constant function that always returns the intervened value. Denoting by $P_{\mcG_i}$ the probability calculated in a model with the causal graph $\mcG_i$, and taking into account for instance the case of $\mcG_1$, that depicts the post-interventional model after forcing $I=yes$, the corresponding interventional query is $P(A_{I=yes}) = P_{\mcG_1}(A | I=yes)$.

Counterfactual queries can be computed using an extended model called the \textit{counterfactual model} (also known as the \textit{twin model}). The twin model is an SCM that includes endogenous variables from both the real and hypothetical scenarios. This is achieved by duplicating the subgraph composed of the endogenous nodes for the real scenario and then applying the intervention.

In the counterfactual model, the endogenous nodes in both the real and hypothetical scenarios share the same exogenous parents, except for the intervened variables. In Figure~\ref{fig:example_cgraph}, $\mcG_2$, $\mcG_3$, and $\mcG_4$ are the graphs corresponding to the twin models for various counterfactual queries in the running example. 

In $\mcG_2,\mcG_3$ and $\mcG_4$, variables from the real scenario are $M$,$I$ and $A$; variables from the hypothetical scenario in which immigration is always present are denoted with a single apostrophe, i.e. $M'$,$I'$ and $A'$; those from the hypothetical scenario in which immigration is never present are denoted with double apostrophe, i.e. $M''$,$I''$ and $A''$.
With this in mind, the query $\text{PNS}(I,A)$ can be calculated as $P_{\mcG_2}(A'=yes, A''=no)$; $\text{PS}(I,A)$ as $P_{\mcG_3}(A'=yes | I = no, A=no)$ and $\text{PN}(I,A)$ as $P_{\mcG_4}(A''=no | I = yes, A=yes)$.  



\section{Complexity discussion}\label{sec:complexity}

The complexity of an SCM increases compared to the original BN. Each variable gains an additional parent, the corresponding exogenous variable, which increases the maximum in-degree by one. For the learning process, the number of variables doubles: the original endogenous variables and the exogenous ones. However, only the exogenous variables are trained. During inference, the number of variables can triple in the worst case, as the twin graph must also be considered.

The main limitation of the method is related to the exponential growth of the cardinality of $U$, which is given by equation \eqref{eq:omega_U}. 
For example, if all variables are binary ($|\Omega_X| = 2$), with 5 parents, the cardinality becomes $|\Omega_U| = 2^{2^5} = 2^{32}$, resulting in over $4000$ million possible outcomes. This implies that, to store only the probability values with 64-bit numbers, 32 GB would be required. With 6 parents, this grows to $2^{2^6} = 2^{64}$, exceeding 18 quintillion outcomes. Such massive spaces make storing probability tables or computing exact inference infeasible.

\bibliographystyle{elsarticle-harv}

\bibliography{main}

\end{document}